%% file: root.tex
\documentclass[journal,letterpaper]{IEEEtran}
\input{preamble}

\begin{document}

\title{A Propagation Perspective on Recursive Forward Dynamics for Systems with Kinematic Loops}

\author{Matthew Chignoli$^{1}$, Nicholas Adrian$^{2}$, Sangbae Kim$^{1}$, Patrick M. Wensing$^{2}$ 

\thanks{$^{1}$ are with the Department of Mechanical Engineering, Massachusetts Institute of Technology (MIT), Cambridge, MA, USA {\tt\small \{chignoli, sangbae\}@mit.edu}}
\thanks{$^{2}$ are with the Department of Aerospace and Mechanical Engineering, University of Notre Dame, Notre Dame, IN, USA {\tt\small \{nadrian, pwensing\}@nd.edu}}
}



\maketitle

\begin{abstract}
We revisit the concept of constraint embedding as a means for dealing with kinematic loop constraints during dynamics computations for rigid-body systems.
Specifically, we consider the local loop constraints emerging from common actuation sub-mechanisms in modern robotics systems (e.g., geared motors, differential drives, and four-bar mechanisms).
However, rather than develop the concept of constraint embedding from the perspective of graphical analysis, we present a novel analysis of constraint embedding that generalizes the traditional concepts of joint models and motion/force subspaces between individual rigid bodies to generalized joint models and motion/force subspaces between groups of rigid bodies subject to loop constraints.
The generalized concepts are used in a self-contained, articulated-body-based derivation of the constraint-embedding-based recursive algorithm for forward dynamics.
The derivation represents the first assembly method to demonstrate the recursivity of articulated inertia computation in the presence of loop constraints.
We demonstrate the broad applicability of the generalized joint concepts by showing how they also lead to the constraint-embedding-based recursive algorithm for inverse dynamics.
Lastly, we benchmark our open-source implementation in \verb!C++! for the forward dynamics algorithm against a state-of-the-art, non-recursive algorithm.
Our benchmarking validates that constraint embedding outperforms the non-recursive alternative in the case of local kinematic loops.
\end{abstract}


\input{Sections/Introduction}
\input{Sections/Background}
\input{Sections/ClusterJointModels}
\input{Sections/PropagationMethodForRecursiveAlgorithmsUsingConstraintEmbedding}
\input{Sections/RNEA}
\input{Sections/Benchmark}
\input{Sections/Conclusion}

\section*{Acknowledgments}
This work was supported by the National Science Foundation (NSF) Graduate Research Fellowship Program under Grant No. 4000092301 and under NSF award CMMI-2220924 with a subaward to the University of Notre Dame. 
The authors would also like to thank Youngwoo Sim from the University of Illinois Urbana-Champaign for his assistance with CAD files and design questions related to the Tello humanoid robot.

\bibliographystyle{ieeetr}
\bibliography{root}

\appendices
\input{Appendices/ValidatingSpanningTreeIdentities}

\vfill

\end{document}

%% file: preamble.tex
\usepackage{amsmath}
\usepackage{amssymb}
\usepackage{cite}
\usepackage{amsthm}
\usepackage{mathtools}
\usepackage[dvipsnames]{xcolor}
\usepackage{hyperref}
\usepackage{graphicx}
\usepackage{algorithm}
\usepackage{algorithmicx}
\usepackage{algpseudocode}
\usepackage{multirow}
\usepackage{upgreek}
\newtheorem{remark}{Remark}
\usepackage{siunitx}
\usepackage{array}

\newcommand{\E}{\mathbf{E}}

\newcommand{\G}{\mathbf{G}}
\renewcommand{\H}{\mathbf{H}}
\newcommand{\I}{\mathbf{I}}

\newcommand{\K}{\mathbf{K}}

\newcommand{\R}{\mathbf{R}}
\renewcommand{\S}{\mathbf{S}}
\newcommand{\T}{^\top}

\newcommand{\X}{\mathbf{X}}

\newcommand{\Z}{\mathbf{Z}}

\newcommand{\f}{\mathbf{f}}
\newcommand{\g}{\mathbf{g}}

\newcommand{\kk}{\mathbf{k}}

\newcommand{\p}{\mathbf{p}}
\newcommand{\q}{\mathbf{q}}

\renewcommand{\v}{\mathbf{v}}

\newcommand{\y}{\mathbf{y}}
\newcommand{\z}{\mathbf{z}}

\newcommand{\aggregateArticle}{a}
\newcommand{\aggregate}{cluster}
\newcommand{\aggregated}{clustered}
\newcommand{\aggregation}{clustering}
\newcommand{\aggregateLink}{cluster}
\newcommand{\Aggregate}{Cluster}
\newcommand{\AggregateLink}{Cluster}
\newcommand{\Aggregating}{Clustering}
\newcommand{\Aggregation}{Clustering}
\newcommand{\aggregateSubscript}{\mathcal{C}}
\newcommand{\agglink}[1]{\mathcal{C}_{#1}}

\newcommand{\cgraph}{\mathcal{G}}
\newcommand{\spantree}{\cgraph_t}

\newcommand{\parentT}[1]{{\lambda_t(#1)}}
\newcommand{\supportT}[1]{\kappa_t\hspace{-1pt}\left(#1\right)}
\newcommand{\childT}[1]{\mu_t\hspace{-1pt}\left(#1\right)}

\newcommand{\aggtree}{\cgraph_\aggregateSubscript}
\newcommand{\parentA}[1]{{\lambda_\aggregateSubscript(#1)}}
\newcommand{\childA}[1]{\mu_\aggregateSubscript\left(#1\right)}
\newcommand{\calJ}{{\mathcal{J}}}

\newcommand{\vcat}[2]{\mathrm{stack}\left(\left\{#2\right\}_{i\in#1}\right)}
\newcommand{\Dcat}[2]{\mathrm{diag}\left(\left\{#2\right\}_{i\in#1}\right)}
\renewcommand{\output}[1]{\Lambda\left(#1\right)}

\newcommand{\EpsilonA}[2]{\epsilon_{#1}(#2)}

\newcommand{\AB}{\mathcal{A}}

\newcommand{\vv}{\boldsymbol{v}}
\newcommand{\vomega}{\boldsymbol{\omega}}
\newcommand{\vn}{\boldsymbol{n}}
\newcommand{\vf}{\boldsymbol{f}}

\newcommand{\motionSpace}{M}
\newcommand{\forceSpace}{F}
\newcommand{\motionSS}{S}
\newcommand{\forceSS}{T}

\newcommand{\gv}{\mathsf{v}}
\newcommand{\gvJ}{\mathsf{v}^\calJ}
\newcommand{\ga}{\mathsf{a}}
\newcommand{\gaJ}{\mathsf{a}^\calJ}

\newcommand{\gf}{\mathsf{f}}
\newcommand{\gfj}{\gf^J}
\newcommand{\gfJ}{\gf^\calJ}
\newcommand{\gfnet}{\gf^\mathrm{net}}

\newcommand{\gtau}{\uptau^\mathrm{tree}}
\newcommand{\gtauInd}{\uptau}

\newcommand{\gCnstrForceInd}{\mathsf{\lambda}^{c}}

\newcommand{\gI}{\mathsf{I}}
\newcommand{\gIA}{\gI^A}
\newcommand{\gIa}{\gI^a}
\newcommand{\gp}{\mathsf{p}}
\newcommand{\gpA}{\gp^A}
\newcommand{\gpa}{\gp^a}

\newcommand{\gD}{\mathsf{D}}

\newcommand{\gu}{\mathsf{u}}
\newcommand{\gU}{\mathsf{U}}
\newcommand{\gL}{\mathsf{L}}

\newcommand{\gS}{\mathsf{S}}
\newcommand{\gSd}{\dot{\gS}}

\newcommand{\gforceSSM}{\mathsf{T}}
\newcommand{\gP}{\mathsf{P}}

\newcommand{\gX}{\mathsf{X}}
\newcommand{\gXM}[2]{{}^{#1}\gX_{#2}}
\newcommand{\gXF}[2]{{}^{#1}\gX_{#2}^*}
\newcommand{\gSKO}[1]{\mathcal{E}_{#1}}
\newcommand{\gSPO}[1]{{}^{\calJ_{#1}}\mathsf{X}_{J_{#1}}}
\newcommand{\gSPOF}[1]{{}^{\calJ_{#1}}\mathsf{X}^*_{J_{#1}}}

\newcommand{\gq}{\mathsf{q}}
\newcommand{\gqd}{\dot{\gq}}
\newcommand{\gqdd}{\ddot{\gq}}

\newcommand{\gy}{\mathsf{y}}
\newcommand{\gyd}{\dot{\mathsf{y}}}
\newcommand{\gydd}{\ddot{\mathsf{y}}}

\newcommand{\gG}{\mathsf{G}}
\newcommand{\gGd}{\dot{\mathsf{G}}}
\renewcommand{\gg}{\mathsf{g}}
\newcommand{\gK}{\mathsf{K}}
\newcommand{\gk}{\mathsf{k}}

\newcommand{\fj}{\f^J}
\newcommand{\fJ}{\f^\calJ}
\newcommand{\fnet}{\f^\mathrm{net}}

\newcommand{\qd}{\dot{\q}}
\newcommand{\qdd}{\ddot{\q}}

\newcommand{\yd}{\dot{\y}}
\newcommand{\ydd}{\ddot{\y}}

\newcommand{\ggamma}{\boldsymbol{\gamma}}
\newcommand{\pphi}{\boldsymbol{\phi}}

\newcommand{\IA}{\I^A}
\newcommand{\invIA}{\boldsymbol{\Phi}^A}
\newcommand{\pA}{\p^A}
\newcommand{\bA}{\mathbf{b}^A}

\newcommand{\Sd}{\dot{\S}}
\newcommand{\Sring}{\mathring{\S}}
\newcommand{\forceSSM}{\mathbf{T}}

\newcommand{\XM}[2]{{}^{#1}\X_{#2}}
\newcommand{\XF}[2]{{}^{#1}\X_{#2}^*}

\newcommand{\ttau}{\boldsymbol{\tau}^\mathrm{tree}}

\newcommand{\cnstrForce}{\boldsymbol{\lambda}^{c,\mathrm{tree}}}

\newcommand{\crm}{\times}
\newcommand{\crf}{\times^*}

\renewcommand{\R}{\mathbb{R}}

\usepackage[nonumberlist, nogroupskip, section=section, numberedsection=autolabel]{glossaries}
\glsdisablehyper

\newacronym{sva}{SVA}{Spatial Vector Algebra}
\newacronym{soa}{SOA}{Spatial Operator Algebra}
\newacronym{rnea}{RNEA}{Recursive Newton-Euler Algorithm}
\newacronym{aba}{ABA}{Articulated-Body Algorithm}

%% file: Sections/Introduction.tex
\section{Introduction}
\begin{figure}
    \centering
    \includegraphics[width=0.85\columnwidth]{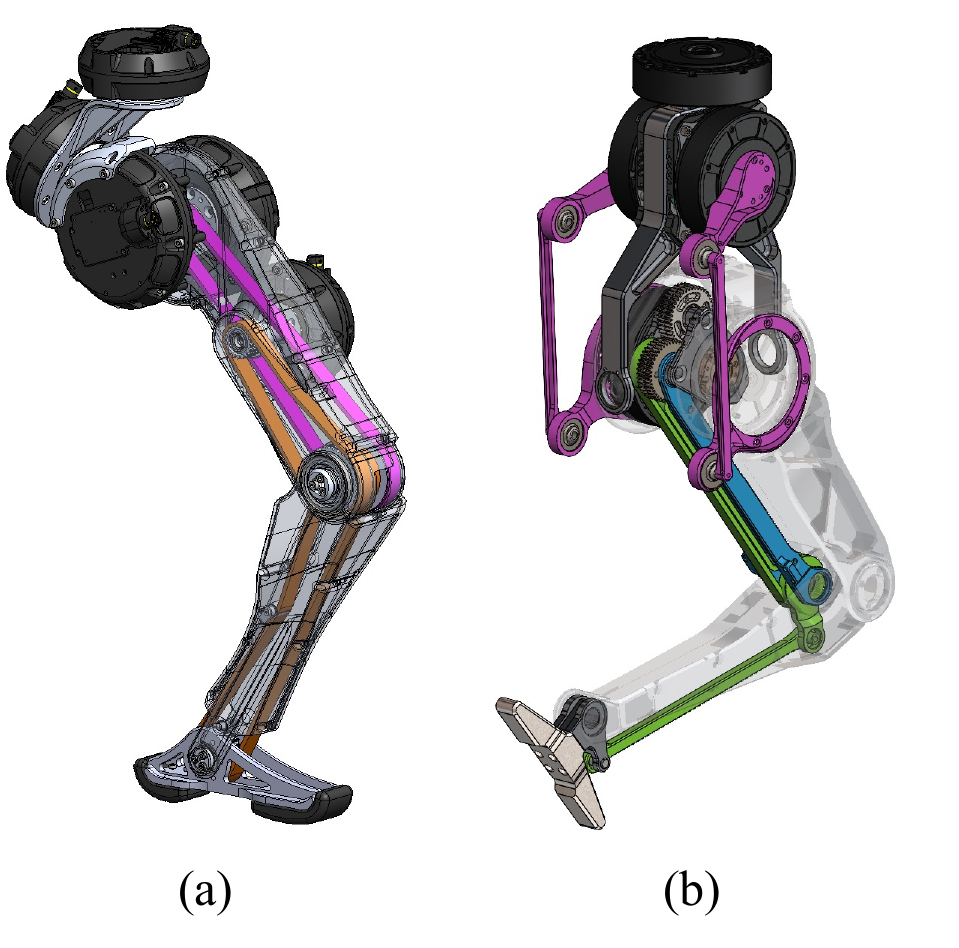}
    \caption{\textbf{(a)} A parallel belt transmission that actuates the knee (pink) and ankle (orange) of the MIT Humanoid's leg \cite{chignoli2021humanoid}. \textbf{(b)} A pair of differential drives that actuate the hip (pink) as well as the knee and ankle (blue and green) of the Tello Humanoid's leg \cite{sim2022tello}.}
    \label{fig:sub_mechanisms}
\end{figure}

In recent years, the field of legged robots has seen a strong trend toward using quasi-direct drives for actuating limbs.
The high torque density, low mechanical impedance, and high bandwidth of such actuators make them appealing for dynamic, contact-rich applications.
A popular quadruped robot design uses a serial architecture, with each leg having three joints and geared motors localized to the respective links (e.g., the AnyMal~\cite{hutter2016anymal}, Mini Cheetah~\cite{katz2019mini}, and A1~\cite{a12023} quadruped robots).
As humanoid robots gain popularity, designs featuring more complex transmission mechanisms across multiple joints have emerged.
For example, the parallel belt transmission of the MIT Humanoid~\cite{chignoli2021humanoid} (Fig.~\ref{fig:sub_mechanisms}a), the differential drives of BRUCE~\cite{liu2022design} and Tello~\cite{sim2022tello} (Fig.~\ref{fig:sub_mechanisms}b), and the linkage mechanisms of Digit~\cite{digit2024} and Kangaroo~\cite{kangaroo2024} enable quasi-direct drives to deliver power to distal joints.

Fast, accurate algorithms are crucial for both model-based control and "model-free" reinforcement learning, which greatly depend on the quality and speed of the underlying simulator.
While the \gls{aba}~\cite{featherstone1983calculation} has traditionally served the role of a fast, accurate algorithm for simulating open-chain robotic systems, it is not suitable for systems with actuation sub-mechanisms like those mentioned above due to the presence of kinematic loops.
Efficient algorithms based on the concept of constraint embedding~\cite{jain2009recursive} have been developed and applied to actuation sub-mechanisms~\cite{kumar2022modular}.
However, despite their elegance, they have yet to proliferate broadly in the field of robotics.
In this paper, we provide a novel perspective on constraint embedding, focusing on how it emerges from the physical analysis of interactions between individual rigid bodies.
The analysis yields a generalized concept of a joint model that can be used to propagate solutions to the loop-constrained dynamics equations through the system.

\subsection{Related Work}

\newcommand{\Ver}{Vereshchagin}
Precursors to the modern \gls{rnea} \cite{Luh1980} and \gls{aba} were developed in the 1970's by Vukobratovic and colleagues~\cite{stepanenko1976dynamics,orin1979kinematic} and \Ver~\cite{vereshchagin1974computer}, respectively.
These algorithms represented a significant improvement relative to the algorithms of their time because they were the first to demonstrate linear complexity in the number of joints of the robot.
\Ver's algorithm, however, did not experience widespread adoption until a decade later, when Featherstone independently developed a closely related recursive forward dynamics algorithm~\cite{featherstone1983calculation} via propagation methods.
The subsequent development of \gls{soa} using graph techniques introduced a systematic way to formulate equations of motion for rigid-body systems and showed how recursive algorithms such as the \gls{aba} can be obtained via operator factorizations of the mass matrix~\cite{rodriguez1987kalman,rodriguez1992spatial}.
By contrast, Featherstone's original development~\cite{featherstone2014rigid,featherstone1983calculation} provided a physical analysis wherein the solution of the Newton-Euler equations for single bodies is propagated to dynamics equations for successively larger submechanisms, and the propagation of those solutions is described by joint models.

At first, recursive algorithms such as the \gls{aba} were only applicable to open chains.
Since then, efforts have been made to generalize the algorithms to capture the effects of loop constraints.
Recognizing the importance of including the complete effects of gear ratios and the gyroscopic effects of the spinning motors, researchers working with high gear-ratio manipulators developed minimally modified versions of classic algorithms such as the \gls{rnea}~\cite{sciavicco1995lagrange,lynch2017modern}, Coriolis matrix factorization algorithm~\cite{becke2012extended}, and \gls{aba}~\cite{murphy1990recursive,jain1990recursive} so that they could handle this particular class of sub-mechanisms. 
The subsequent development of general assembly algorithms such as the Divide-and-Conquer Algorithm~\cite{featherstone1999divide1,featherstone1999divide2} and the Assembly-Disassembly Algorithm~\cite{yamane2009comparative} used the concepts of multi-handle articulated bodies and inverse articulated inertia to compute the forward dynamics of systems with a larger class of loop constraints.
While these algorithms excel when parallel computing is available, they are not optimized for serial performance in the way that \gls{aba} is.

Alternatively,~\cite{sathya2023efficient} develops a family of constrained dynamics algorithms using a Gauss Principle of Least Constraint-based approach similar to~\cite{popov1978manipuljacionnyje}.
Their method is specialized to handle endpoint constraints, which are different than the ``internal" loop constraints such as those resulting from actuation sub-mechanisms as considered herein. 
Efficient non-recursive algorithms for constrained forward dynamics of systems with arbitrary closure constraints were also developed.
These algorithms use techniques such as sparse matrix factorization~\cite{featherstone2005efficient,carpentier2021proximal} to efficiently solve the large systems of linear equations arising from the equations of motion and closure constraints. A recursive counterpart to these methods was proposed by Brandl et al., which considers Lagrange multiplier propagation/elimination~\cite{brandl1987algorithm} to address loop constraints sequentially rather than for the system as a whole.

A critical development in constrained forward dynamics was manipulating systems such that, even in the presence of kinematic loops, they would be topologically compatible with conventional recursive algorithms for open chains.
The first implementation of this idea was called Recursive Coordinate Reduction ~\cite{critchley2003generalized} and involved introducing ``phantom bodies" into the system to sever the closed loops artificially.
Jain's development of constraint embedding~\cite{jain2009recursive} achieved the same effect without introducing phantom bodies.
Instead, constraint embedding involves grouping bodies involved in a kinematic loop into an ``aggregate link" and then using the topology of the aggregated open-chain structure that emerges to formulate recursive algorithms mirroring those for conventional open-chains.
Recent use cases for constraint embedding have demonstrated its utility in efficiently resolving the dynamics of series-parallel hybrid robots~\cite{muller2022constraint,kumar2022modular}.

\begin{figure}
    \centering
    \includegraphics[width=\columnwidth]{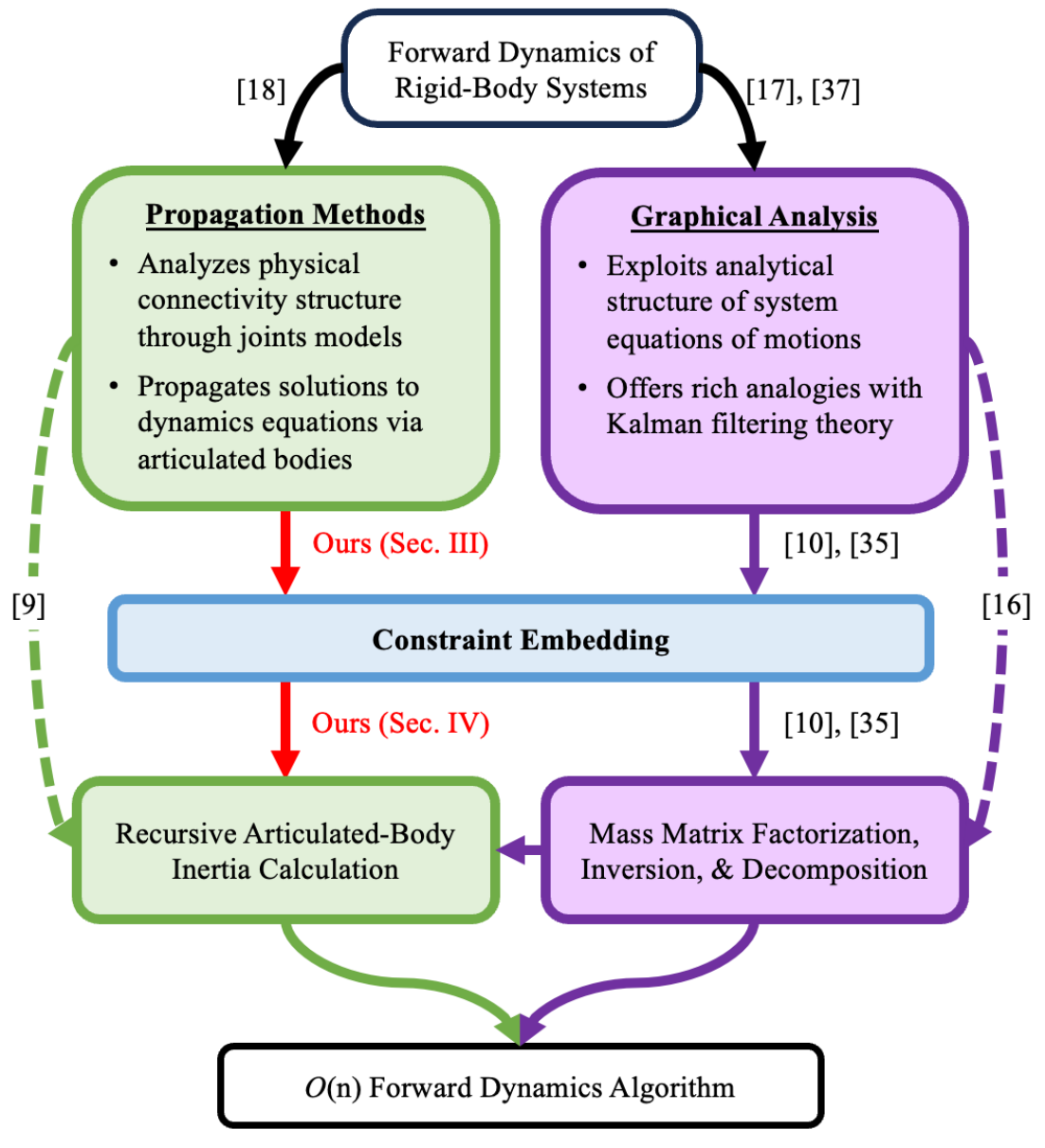}
    \caption{Illustration of two common approaches to analyzing rigid-body systems and how they lead to the same efficient algorithm for computing forward dynamics. Dashed lines represent connections that are only valid for robots without loop constraints.}
    \label{fig:contribution}
\end{figure}

\subsection{Contribution}
This paper provides a novel perspective on constraint embedding and its corresponding algorithmic developments that emphasizes connections to conventional physical analysis involving joint models, motion and force subspaces, and articulated bodies~\cite{featherstone2014rigid}.
In this sense, we view our work as complementary to the multibody graph transformation perspective of \gls{soa}~\cite{jain2012multibody_part1,jain2012multibody_part2}, which developed constraint embedding via partitioning, aggregation, and substructuring of Spatial Kernel Operator matrices.
We contextualize our contribution and illustrate its complementarity to the original development of constraint embedding in Fig.~\ref{fig:contribution}.
The specific contributions of our work are:
\begin{itemize}
    \item Develop a generalized concept of joint models between aggregate links through appeals to traditional joint model analysis that also consider the effects of loop constraints,
    \item Use the generalized joint model in an articulated-body assembly process leading to a novel, propagation-method-based derivation of the constraint-embedding \gls{aba},
    \item Demonstrate the utility of the generalized joint model by applying it to constrained inverse dynamics.
\end{itemize}
While the original constraint-embedding \gls{aba} has been reinterpreted using Featherstone's numbering convention and Lie Group concepts in place of SOA conventions~\cite{kumar2022modular}, our work analyzes the physical meaning of generalized joint quantities (e.g., the force transmitted across a generalized joint) that serve our subsequent algorithm derivations.
Furthermore, our novel multi-handle articulated-body assembly process is the first assembly process to demonstrate the recursivity of articulated inertia computation in the presence of loop constraints.

We provide an open-source C++ implementation of the library of algorithms at~\url{https://github.com/ROAM-Lab-ND/generalized_rbda}.
To demonstrate the situations in which constraint embedding is especially efficient, we benchmark our library's constrained forward dynamics algorithm against a state-of-the-art, non-recursive algorithm from the Pinocchio library~\cite{carpentier2019pinocchio}.
The benchmark considers a range of robots that vary in their number of bodies, connectivity, types of constraints, and locations of constraints.

The rest of this paper will be organized as follows. 
Section~\ref{sec:background} provides background info on modeling rigid-body systems.
The concept of a generalized joint model for constraint-embedded systems is developed in Section~\ref{sec:agg_joint_model}.
The propagation-method derivation of the constraint-embedding \gls{aba} is developed in Section~\ref{sec:cluster-propagation}, and  Section~\ref{sec:rnea_connection} uses the generalized joint model to derive the constraint-embedding RNEA.
The constraint-embedding algorithms in our open-source repository are benchmarked against the state-of-the-art, non-recursive algorithms in Pinocchio in Section~\ref{sec:benchmark}.
Section~\ref{sec:conclusion} summarizes and discusses the broader impacts of the work. 

%% file: Sections/Background.tex
\section{Background} \label{sec:background}

\subsection{Modeling} \label{ssec:modeling}

A popular approach to modeling rigid-body systems is to describe the system in terms of its component parts via a \textit{system model}.
A system model consists solely of rigid bodies and the kinematic relationships between them (i.e., joints).
The system model is often represented with a graph~\cite{featherstone2014rigid, jain2011graph1}.
In this work, we use the \textit{connectivity graph} convention from~\cite{featherstone2014rigid}:
\begin{itemize}
    \item Its nodes represent bodies.
    \item Its edges represent joints.
    \item Exactly one node represents a fixed base.
    \item The graph is undirected.
    \item The graph is connected.
\end{itemize}

\begin{figure}
    \centering
    \includegraphics[width=\columnwidth]{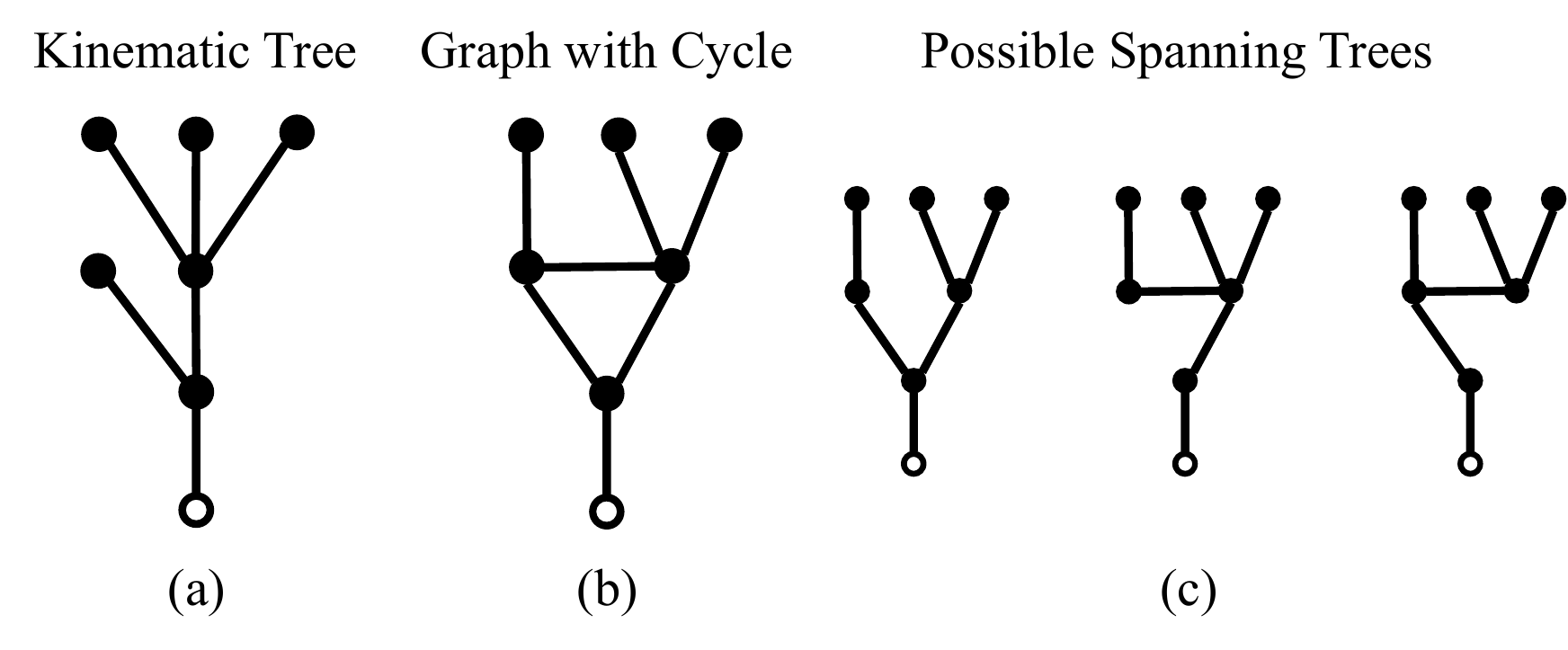}
    \caption{\textbf{(a)} Example of a connectivity graph that is a kinematic tree. \textbf{(b)} Example of a connectivity graph that has a cycle and is therefore not a kinematic tree. \textbf{(c)} Every possible spanning tree for the example in (b).}
    \label{fig:rbcg}
\end{figure}

A graph is a topological tree if there exists only one path between any two nodes in the graph (i.e., no cycles).
A rigid-body system is called a kinematic tree when its connectivity graph is a topological tree.
For a connectivity graph $\cgraph$, a spanning tree $\spantree$ is any subgraph of $\cgraph$ that contains all nodes in $\cgraph$ along with a subset of edges such that $\spantree$ is a topological tree. 
The relationship between connectivity graphs, kinematic trees, and spanning trees is visualized in Fig.~\ref{fig:rbcg}.

Finding at least one spanning tree for a connectivity graph is always possible.
If the number of edges in the connectivity graph is greater than or equal to the number of nodes in the connectivity graph, there will be one or more edges in the connectivity graph that are not in the spanning tree.
These non-tree edges are called chords.
Joints associated with edges in the spanning tree are called \textit{tree joints}, while joints associated with chords are called \textit{loop joints}.

We use the ``regular numbering" scheme from~\cite{featherstone2014rigid} to identify the nodes and edges (bodies and joints) of the connectivity graph.
An example of regular numbering is shown in Fig.~\ref{fig:aggregated_kinematic_tree}b.
To describe the relationships between elements in the spanning tree, we use the following definitions adapted from~\cite{featherstone2014rigid}:
\begin{itemize}
    \item $\parentT{i}$: the body number of the parent of body $i$ in $\spantree$,
    \item $\supportT{i}$: the set of numbers corresponding to the supporting bodies of body $i$ in $\spantree$, defined by 
    \[
    \supportT{i} = i \cup \supportT{\parentT{i}}, \text{ where } \supportT{0} = \emptyset.
    \]
    \item $\childT{i}$: the set of body numbers of the children of body $i$ in $\spantree$, defined by 
    \[
    {\childT{i} = \{j : \parentT{j} = i\}}.
    \]
    Bodies with no children, i.e., bodies where $\childT{i}$ is empty, are referred to as ``leaves" of the tree.
\end{itemize}

\subsection{Spatial Vector Algebra} \label{ssec:sva}
We assume readers have a working knowledge of conventional $6$D Spatial Vector Algebra.
Thus, we provide a brief review here and refer readers to the \gls{sva} tutorials~\cite{featherstone2010beginner,featherstone2010beginner2} for more details.

The foundational pieces of \gls{sva} are the 6D spatial motion and spatial force vectors.
Spatial motion vectors combine the angular and linear components of motion via 
\[
^{O}\v = \begin{bmatrix} ^{O}\vomega \\ ^{O}\vv_O \end{bmatrix},
\]
where the superscript $O$ denotes a quantity expressed in the coordinate frame $O$, $^{O}\v$ is the spatial motion, $^{O}\vomega\in\R^3$ is the angular velocity of the body, and $^{O}\vv_O\in\R^3$ is the linear velocity of a body-fixed point instantaneously coincident with the origin of frame $O$.
Similarly, spatial force vectors combine the linear and angular components of force interactions via
\[
^{O}\f = \begin{bmatrix} ^{O}\vn_O \\ ^{O}\vf \end{bmatrix},
\]
where $^{O}\f$ is the spatial force, $^{O}\vn_O\in\R^3$ is the net moment about the origin of frame $O$, and $^{O}\vf\in\R^3$ is the net linear force on the body.

The frame of expression for spatial vectors can be changed through spatial transforms.
Spatial transforms depend on the relative orientation and position of the coordinate frames that they transform between.
The spatial motion transform $\XM{P}{O}$ provides the coordinate transform ${}^P\v  = \XM{P}{O}{}^O\v$, while the spatial force transform $\XF{P}{O}$ provides the coordinate transform ${}^P\f  = \XF{P}{O}{}^O\f$.
These transforms are related via the following identities: $\XM{O}{P} = (\XM{P}{O})^{-1}$, $\XF{O}{P} = (\XF{P}{O})^{-1}$, and $\XM{O}{P}^\top = \XF{P}{O}$.

\subsection{Loop Constraints}
We begin by considering a rigid-body kinematic tree in the absence of constraints, i.e., for its spanning tree. The equations of motion for such a system are given by
\begin{equation}
\H(\q)\qdd + \mathbf{c}(\q,\qd) = \ttau, \label{eqn:uncon_eom}
\end{equation}
where $\q$ is the set of spanning tree coordinates, $\H$ is the mass matrix, $\mathbf{c}$ is the bias force, and $\ttau$ is the set of generalized active forces associated with the spanning tree coordinates.
Beyond~\eqref{eqn:uncon_eom}, systems with loop joints must also consider loop constraint forces and satisfy the kinematic constraints associated with those loop joints.

An independent kinematic loop of a connectivity graph is a cycle that traverses one chord.
The number of independent kinematic loops always equals the number of chords~\cite{featherstone2014rigid}.
The loop joint associated with a chord encodes the kinematic constraint between all the bodies involved in an independent kinematic loop. 
These constraints can be expressed in either ``implicit" form \cite{featherstone2014rigid},
\begin{equation}
    \pphi(\q) = 0, \quad \K(\q) \qd = 0, \quad \K(\q)\qdd=\kk(\q,\qd),
\end{equation}
or, in cases where an independent set of coordinates exists, explicit form~\cite{featherstone2014rigid}
\begin{equation}
\q = \ggamma(\y), \quad \qd = \G(\q) \yd, \quad \qdd = \G(\q)\ydd + \g(\q,\qd), \label{eqn:explicit_constraints}
\end{equation}
where $\q$ is the set of spanning coordinates of the robot and $\y$ is the set of independent coordinates ($\mathrm{dim}~\y \le \mathrm{dim}~\q$).
The constraint Jacobians and biases for the implicit and explicit constraints are given by
\begin{equation}
\K = \frac{\partial \pphi(\q)}{\partial \q}, \quad \kk = -\dot{\K}\qd, \quad \G = \frac{\partial\ggamma(\y)}{\partial\y}, \quad \g = \dot{\G}\yd. \label{eqn:constraint_jacobians}
\end{equation}
We treat all constraints in this work as explicit.
However, as we discuss later in Sec.~\ref{sec:cluster-propagation}, the final algorithm does not rely on a definition of $\ggamma(\y)$, and so only depends on the adoption of independent velocities and the formulation of explicit constraint Jacobians and biases, $\G$ and $\g$.
Such quantities can be numerically extracted from any implicit constraint following the approach in~\cite{kumar2022modular}.

\begin{figure}
    \centering
    \includegraphics[width=\columnwidth]{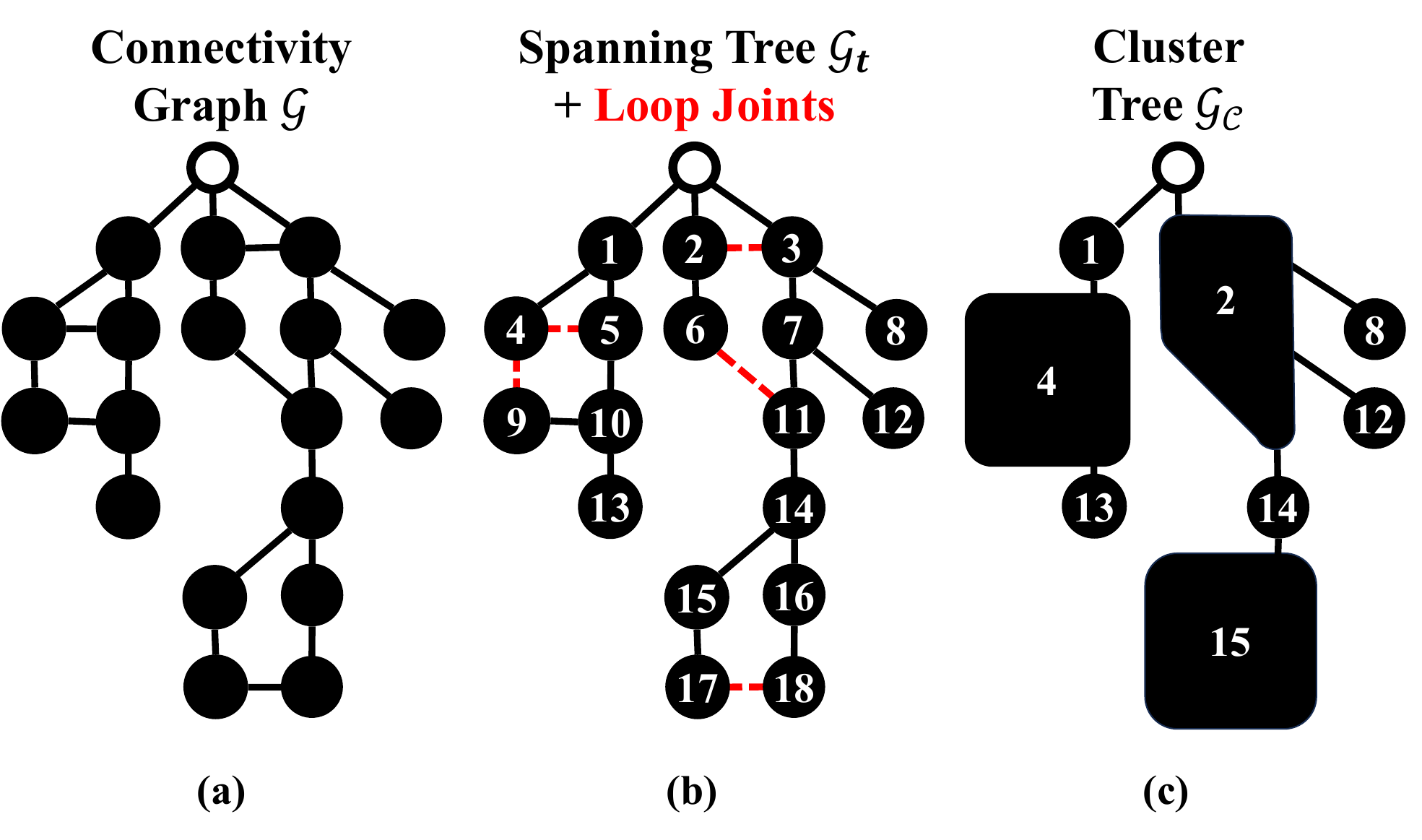}
    \caption{\textbf{(a)} A connectivity graph containing cycles. \textbf{(b)} A regularly numbered spanning tree for the connectivity graph. The spanning tree contains nested loops (chord between nodes 2 and 3 is nested by the chord between nodes 6 and 7) and parallel loops where a single body is involved in multiple loop constraints (node 4). \textbf{(c)} The \aggregate~tree that emerges from applying constraint embedding to (b).}
    \label{fig:aggregated_kinematic_tree}
\end{figure}

\subsection{Multibody Graph Transformations}

Constraint embedding allows \textit{any} connectivity graph to be converted to a related graph with tree topology~\cite{jain2009recursive}.
The technique works by isolating sub-groups of bodies involved in loop constraints.
The internal kinematic/dynamics relationships between these bodies in a sub-group must be defined, and so, too, must the coupling of the sub-group to the rest of the system.
These internal and coupling relationships were originally presented via substructuring of the Spatial Kernel Operator model matrices~\cite{jain2012multibody_part2}.
In this work, we adopt a propagation perspective to rigid-body dynamics that seeks to describe the propagation of solutions to dynamics equations through the use of joint models that capture these internal and coupling relationships.

An isolated sub-group can be represented as a single node in the new ``\aggregate~connectivity graph."
The sub-group of bodies represented by a single node in the graph is called \aggregateArticle~``\aggregateLink" herein.\footnote{These sub-groups of bodies are called ``aggregate links" in the \gls{soa} literature~\cite{jain2009recursive}.}
\Aggregating~all sub-groups involved in loop constraints into their respective \aggregateLink s will always lead to \aggregateArticle~``\aggregate~tree" $\aggtree$ that contains all of the bodies from the original connectivity graph but grouped such that every node in $\aggtree$ has a single parent~\cite{jain2012multibody_part2}.
An example of a connectivity graph containing loop joints being converted to \aggregateArticle~\aggregate~tree is shown in Fig.~\ref{fig:aggregated_kinematic_tree}.
The \aggregation~method of~\cite{jain2012multibody_part2} deals with the cases of nested and parallel loop constraints as shown in Fig.~\ref{fig:aggregated_kinematic_tree}c, as well as any other possible cases, by grouping all bodies into a single \aggregateLink, which is itself subject to the effects of all loop constraints between its constituent bodies.
The \aggregate~tree is topologically compatible with recursive algorithms for kinematic trees such as the \gls{rnea} and \gls{aba}.

Complete details on optimally carrying out constraint-embedding \aggregation~are given in~\cite{jain2012multibody_part2}.
Optimal \aggregation, in this case, is the \aggregation~that leads to \aggregateLink s containing the fewest numbers of bodies.
This is because the complexity of the internal kinematics/dynamics of the \aggregateLink s grows as the number of contained bodies grows, and so there reaches a point where using constraint embedding to deal with constraints leading to large \aggregateLink s becomes less computationally efficient than dealing with those constraints using non-recursive alternatives~\cite{jain2012efficient}.
Sec.~\ref{sec:benchmark} investigates this point.

Lastly, we define the numbering scheme and connectivity definitions to describe the elements of $\aggtree$.
For ease of notation, we will treat all nodes in $\aggtree$ as \aggregateLink s.
Rigid bodies not involved in loop constraints will be treated as \aggregateLink s containing a single body with no associated loop constraints. 
This adds no algorithmic or computational complexity to any subsequent derivations.
\AggregateLink s in $\aggtree$ are identified by the body number of their lowest-numbered constituent body.
For example, in Fig.~\ref{fig:aggregated_kinematic_tree} the \aggregateLink~containing bodies 2, 3, 6 and 7 will be identified as \aggregateLink~2.
We use the set $\agglink{i}$ to represent the body numbers corresponding to the bodies in the \aggregateLink~$i$.
Individual bodies in the \aggregateLink~can be identified via $\agglink{i}(j)$, which gives the $j$th body number in the sequence of body numbers for \aggregateLink~$i$, ordered from lowest to highest. 
The definitions for $\parentA{\cdot}$ and $\childA{\cdot}$ are identical to those in Sec.~\ref{ssec:modeling}, but refer to connectivity in $\aggtree$ rather than $\spantree$.
The base bodies in \aggregateArticle~\aggregateLink~are defined as those whose parent bodies are not in that \aggregateLink.
For example, the base bodies of \aggregateLink~15 in Fig.~\ref{fig:aggregated_kinematic_tree}c are bodies $15$ and $16$.
Finally, the output body of \aggregateArticle~\aggregateLink~is the parent body for all of the base bodies in that \aggregateLink.\footnote{Subsequent derivations can be extended to a more general case where instead of requiring $\parentT{i}=\output{k}$ for all base bodies $i$ in \aggregateLink~$k$, it is only required that $\parentT{i}\in\parentA{k}$ for all base bodies $i$ in \aggregateLink~$k$.
However, because our assumption covers all of the sub-mechanisms we are interested in, this extension allowing for multiple output bodies is left to future work.}
We use $\output{k}$ to denote the output body of \aggregateLink~$k$;
for example, in Fig.~\ref{fig:aggregated_kinematic_tree}, $\output{4}=1$ and $\output{14}=11$.

%% file: Sections/ClusterJointModels.tex
\section{\Aggregate~Joint Model} \label{sec:agg_joint_model}

We now discuss the kinematic/dynamic relationships within \aggregateLink s and their coupling with the rest of the system.
We do so by analyzing the interactions between individual rigid bodies under the influence of loop constraints to generalize the concept of traditional joint constraints to \aggregate~joints.
To that end, we introduce a new term related to constraint embedding: \aggregate~joint.
The \aggregate~joint connecting \aggregateArticle~\aggregateLink~to its parent in $\aggtree$ includes the loop joints relating the bodies within the \aggregateLink~{\em and} the tree joints connecting bodies in the \aggregateLink~to their parent bodies in $\spantree$.
We differentiate \aggregate~joints from individual tree joints with the notation $\calJ$ and $J$, respectively. 

In our analysis of interactions between individual rigid bodies, we arrive at the same mathematical joint structure as in Sec. 3 of~\cite{jain2012multibody_part2}, but from a new, complementary perspective.
Rather than substructuring Spatial Kernel Operator matrices,
we develop the generalized of a joint model through appeals to traditional tree joint model analysis that also consider the influence of loop constraints.
This new perspective also provides the foundation for the propagation-method-based derivation of recursive forward dynamics in Sec.~\ref{sec:cluster-propagation}.

\subsection{\Aggregation~Operators}

The \gls{sva} quantities used to describe individual rigid bodies can easily be generalized to describe \aggregateLink s simply by concatenating the \gls{sva} quantities of \aggregateArticle~\aggregateLink's constituent bodies.
We represent this concatenation by the following operators: ``$\vcat{\agglink{k}}{\z_i}$" creates a single vector containing the vector quantities $\z$ corresponding to every rigid body in \aggregateLink~$k$ and  ``$\Dcat{\agglink{k}}{\Z_i}$" creates a block diagonal matrix of the matrix quantities $\Z$ corresponding to every rigid body in \aggregateLink~$k$.

For example, the spanning joint state for \aggregateArticle~\aggregate~joint is given by
\begin{equation}
    \gq_k \triangleq \vcat{\agglink{k}}{\q_i}, \quad \gtau_k=\vcat{\agglink{k}}{\ttau_k}, \label{eqn:agg_q}
\end{equation}
where $\q_i$ and $\ttau_i$ are the joint coordinate and generalized active force for the $i$th tree joint in $\spantree$, respectively.
We further define $\gG_k$ as the matrix capturing all loop constraints within \aggregate~joint $k$ so that
\begin{equation}
    \gqd_k \triangleq \gG_k\gyd_k \label{eqn:agg_y}
\end{equation}
relates the independent coordinates $\gy_k$ for \aggregate~joint $k$ to those of the spanning tree.
To simplify later developments, we assume $\gy_k$ to be a subset of spanning tree joint coordinates and that these coordinates coincide with the actuated joints.
In this most common case, $\gtauInd_k$ coincides with the vector of actuator efforts for the~\aggregateLink.%
\footnote{\label{foot:general_actuation} More generally, for some vector of actuator efforts $\overline{\uptau}_k$ for the \aggregateLink, there always exists a matrix $\mathsf{W}_k(\gq_k)$ so that the power delivered by the efforts is $\gyd_k\T \mathsf{W}_k\T \overline{\uptau}_k$. The generalized force associated with $\gy_k$ is then $\gtauInd_k = \mathsf{W}_k\T \overline{\uptau}_k$. This construction enables treating, for example, overactuated clusters or situations where loop joints are actuated. } 

\subsection{\Aggregate~Motion Subspace} \label{ssec:agg_motion_ss}

Unlike a traditional joint that imposes constraints on the relative velocity between two rigid bodies, \aggregateArticle~\aggregate~joint imposes constraints on the relative velocity between the bodies in \aggregateArticle~\aggregateLink~and the \aggregateLink 's output body.

The relative velocity between a body in \aggregateLink~$k$ and its output body is given by
\begin{equation}
    \v^\calJ_i \triangleq \v_i - {}^i\v_{\output{k}}.
\end{equation}
Here, and for the remainder of the paper, we omit the preceding superscript when a quantity is expressed in its local frame.
In other words, $\z_i\triangleq{}^i\z_i$ for any spatial quantity $\z$.

Because $\v_i$ cannot be computed in isolation without also computing the velocities of all other bodies in the \aggregateLink, it is more useful to describe the relative velocity across \aggregateArticle~\aggregate~joint as a stacked quantity
\begin{equation}
    \gvJ_k \triangleq \vcat{\agglink{k}}{\v^\calJ_i} = \vcat{\agglink{k}}{\v_i - {}^i\v_{\output{k}}}. \label{eqn:define_vJ}
\end{equation}
For this work, we assume that no joint constraints depend on time.
Therefore, the relative velocities $\gvJ_k$ are restricted to a configuration-dependent subspace $\motionSS_k\subseteq\motionSpace^{6n_b}$, where $n_b$ is the number of bodies in \aggregateLink~$k$ and $\motionSpace^{6n_b}$ is the $6n_b$-dimensional space of stacked spatial motion vectors~\cite{featherstone2014rigid}.
This subspace associated with the \aggregate~joint is determined by both the tree and the loop joints contained by the \aggregate~joint.
It is encoded by the \aggregate~motion subspace matrix $\gS_k$, such that
\begin{equation}
    \gvJ_k \triangleq \gS_k(\gq_k)\gyd_k. \label{eqn:vagg_J}
\end{equation}
We remove the explicit dependence of $\gS_k$ on $\gq_k$ in subsequent developments for notational conciseness.

Including the effects of the loop joints is the critical difference between \aggregateArticle~\aggregate~motion subspace and a conventional motion subspace.
Consequently, there is no straightforward mapping from ``\aggregate~joint type" to \aggregate~motion subspace matrix as there is for conventional tree joints.
The motion subspace matrix for a conventional tree joint encodes the relative velocity allowed between a body and its parent,
\begin{equation}
    \v^J_i \triangleq \v_i - \v_\parentT{i} \triangleq \S_i\qd_i. \label{eqn:tree_joint_mssm}
\end{equation}
For example, an individual revolute joint about the $z$-axis has the motion subspace matrix $\mathbf{S}=\begin{bmatrix} 0 & 0 & 1 & 0 & 0 & 0 \end{bmatrix}^\top$, and an individual prismatic joint along the $y$-axis has the motion subspace matrix $\mathbf{S}=\begin{bmatrix} 0 & 0 & 0 & 0 & 1 & 0 \end{bmatrix}^\top$.
For \aggregateArticle~\aggregate~joint, since the details are dependent on the loop closures, it is much more difficult to identify canonical cases that will commonly appear.

However, any \aggregate~motion subspace matrix can be constructed using the motion subspaces from the \aggregateLink 's individual tree joints and loop constraint Jacobians.
We begin this construction by finding a matrix $\gSPO{k}$ that relates the relative velocities across the individual tree joints to relative velocities across the \aggregate~joint.
This matrix $\gSPO{k}$ is comprised of spatial motion transforms between individual rigid bodies and satisfies the tree joint to \aggregate~joint relationship
\begin{equation}
\begin{split}
\vcat{\agglink{k}}{\v_i - {}^i\v_{\output{k}}} &= \\
&\hspace{-2cm}\gSPO{k}\vcat{\agglink{k}}{\v_i - {}^i\v_\parentT{i}}. \label{eqn:motion_par_rel_to_out_rel}
\end{split}
\end{equation}
Each difference $\v_i - {}^i\v_{\output{k}}$ can alternatively be written using differences with respect to spanning tree parents, summed over the bodies in \aggregateLink~$k$ that support body $i$,
\begin{equation}
    \v_i - {}^i\v_{\output{k}} = \sum_{j\in\supportT{i}\cap\agglink{k}} \XM{i}{j}(\v_j - {}^j\v_\parentT{j}).
\end{equation}
Therefore, we can write $\gSPO{k}$ as
\begin{equation}
\gSPO{k} = \begin{bmatrix}
    \mathbf{1} & \mathbf{0} & \hdots & \mathbf{0} \\
    \X_{2,1} & \mathbf{1} &  & \vdots \\
    \vdots &  & \ddots & \mathbf{0} \\
    \X_{n,1} & \hdots & \X_{n,n-1} & \mathbf{1}
\end{bmatrix}, \label{eqn:internal_motion_transform}
\end{equation}
with
\[
\X_{i,j} = \begin{cases}
      \XM{\agglink{k}(i)}{\agglink{k}(j)}, & \text{if}\ \agglink{k}(j)\in\supportT{\agglink{k}(i)} \\
      \mathbf{0}, & \text{otherwise}
    \end{cases}.
\]
Since we follow regular numbering, $\gSPO{k}$ is always block unit lower triangular.
This matrix is equivalent to the Spatial Propagator Operator for the spatial motion transforms $\XM{i}{\parentT{i}}$, originally developed in~\cite{jain2011graph1}.

Proceeding with the construction,
\begin{equation}
\begin{split}
    \gvJ_k = &\hspace{2pt}\vcat{\agglink{k}}{\v_i - {}^i\v_{\output{k}}} \\ 
    \overset{\eqref{eqn:motion_par_rel_to_out_rel}}{=} &\gSPO{k}\vcat{\agglink{k}}{\v_i - {}^i\v_\parentT{i}} \\
    \overset{\eqref{eqn:tree_joint_mssm}}{=} &\gSPO{k}\vcat{\agglink{k}}{\S_i\qd_i} \\
    \overset{\eqref{eqn:agg_q}}{=} &\gSPO{k}\Dcat{\agglink{k}}{\S_i}\gqd_k \\
    \overset{\eqref{eqn:agg_y}}{=} &\gSPO{k}\Dcat{\agglink{k}}{\S_i}\gG_k\gyd_k,
\end{split} \label{eqn:aggregate_motionSSM_construction_process}
\end{equation}
And so, therefore,
\begin{equation}
    \gS_k = \gSPO{k}\Dcat{\agglink{k}}{\S_i}\gG_k. \label{eqn:aggregate_motionSSM_construction}
\end{equation}

The \aggregate~joint's acceleration constraint equations are likewise obtained by propagating the acceleration constraints between individual tree joints to acceleration constraints between bodies in \aggregateArticle~\aggregateLink~and their output body,
\begin{equation}
\begin{split}
\gaJ_k \triangleq \vcat{\agglink{k}}{\mathbf{a}_i - {}^i\mathbf{a}_{\output{k}}} &= \\
&\hspace{-3cm}\gSPO{k}\vcat{\agglink{k}}{\mathbf{a}_i - {}^i\mathbf{a}_\parentT{i}}.
\end{split}
\end{equation}
We use the dot and ring notation from~\cite{featherstone2014rigid} to express the acceleration constraint across an individual tree joint,
\begin{equation}
\begin{split}
    \mathbf{a}_i - {}^i\mathbf{a}_\parentT{i} &= \S_i\qdd_i + \Sd_i\qd_i \\
    &= \S_i\qdd_i + \Sring_i\qd_i + \v_i \crm \S_i\qd_i,
\end{split} \label{eqn:tree_joint_accel}
\end{equation}
where $\Sring_i$ is the ``apparent derivative" of $\S_i$ in a coordinate system that is moving with body $i$.
Performing a similar construction to~\eqref{eqn:aggregate_motionSSM_construction_process} yields,
\begin{equation}
\begin{split}
    \gaJ_k \overset{\eqref{eqn:tree_joint_accel}}{=} &\gSPO{k}\vcat{\agglink{k}}{\S_i\qdd_i + \Sring_i\qd_i + \v_i \crm \S_i\qd_i} \\
    \overset{\eqref{eqn:agg_q}}{=} &\gSPO{k}\Dcat{\agglink{k}}{\S_i}\gqdd_k + \\
    &\hspace{1cm}\gSPO{k}\Dcat{\agglink{k}}{\Sring_i + \v_i \crm \S_i}\gqd_k \\
    \overset{\eqref{eqn:explicit_constraints}}{=} &\gSPO{k}\Dcat{\agglink{k}}{\S_i}\left(\gG_k\gydd_k + \gGd_k\gyd_k\right) + \\
    &\hspace{1cm}\gSPO{k}\Dcat{\agglink{k}}{\Sring_i + \v_i \crm \S_i}\gG_k\gyd_k \\
    \overset{\eqref{eqn:aggregate_motionSSM_construction}}{=} &\gS_k\gydd_k + \gSPO{k}\left(\Dcat{\agglink{k}}{\S_i}\gGd_k\right)\gyd_k + \\
    &\hspace{10pt}\gSPO{k}\Dcat{\agglink{k}}{\Sring_i + \v_i \crm \S_i}\gG_k\gyd_k.
\end{split} \label{eqn:aggregate_accel_cnstr_construction_process}
\end{equation}
If we define
\begin{equation}
\begin{split}
    &\gSd_k \triangleq \gSPO{k}\bigg(\Dcat{\agglink{k}}{\S_i}\gGd_k + \\ &\hspace{2.5cm}\Dcat{\agglink{k}}{\Sring_i + \v_i \crm \S_i}\gG_k\bigg), \label{eqn:aggregate_velocity_product_accel}
\end{split}
\end{equation}
we arrive at an expression for the acceleration constraint across a cluster joint that resembles the constraint across an individual joint~\eqref{eqn:tree_joint_accel},
\begin{equation}
    \gaJ_k = \gS_k\gydd_k + \gSd_k\gyd_k. \label{eqn:agg_aJ}
\end{equation}

\begin{figure*}
    \centering
    \includegraphics[width=2.0\columnwidth]{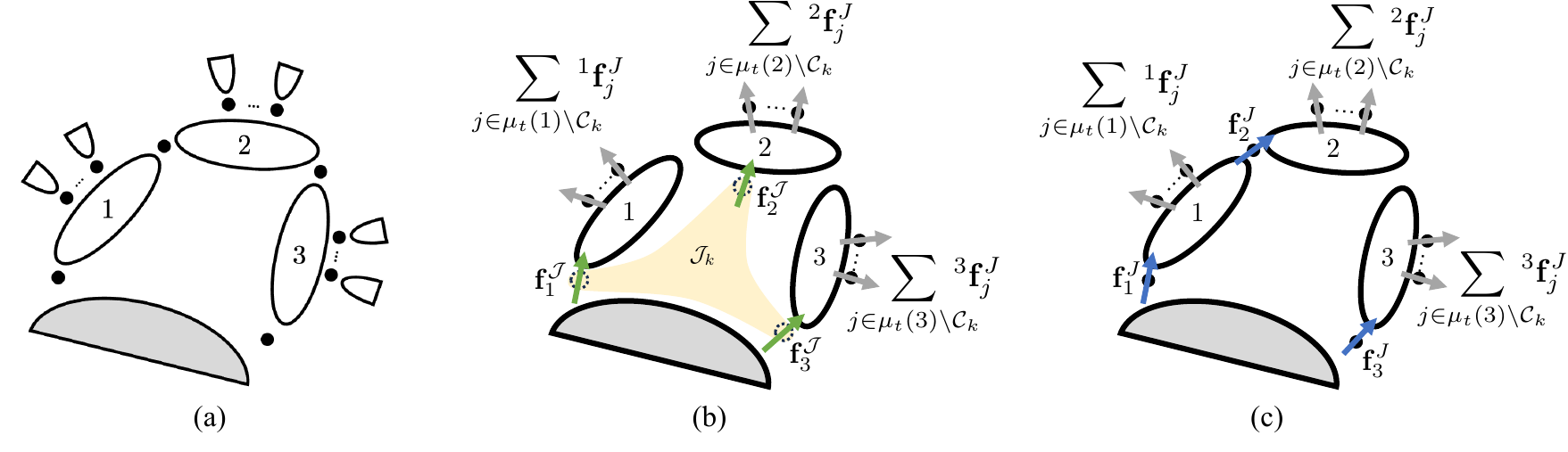}
    \caption{\textbf{(a)} Bodies 1, 2, and 3 are part of a four-bar mechanism resulting in \aggregateLink~$k$. \textbf{(b)} Forces acting on the bodies in \aggregateLink~$k$, expressed in terms of the \aggregate~forces $\fJ_i$ being transmitted across \aggregate~joint $k$ to the bodies in \aggregateLink~$k$. \textbf{(c)} Statically equivalent representation of the forces in (b), expressed using the forces $\fj_i$ transmitted across individual tree joints in the spanning tree.}
    \label{fig:statically_equivalent_forces}
\end{figure*}

\subsection{\Aggregate~Force Subspace} \label{ssec:agg_force_ss}
To restrict the motion of the bodies in the \aggregateLink~to the motion subspace $\motionSS_k$, the \aggregate~joint must exert constraint forces that resist motion in the directions not spanned by $\motionSS_k$.
In other words, the constraint forces of the \aggregate~joint are elements of the orthogonal complement $\motionSS_k^\perp\subseteq\forceSpace^{6n_b}$, where $\forceSpace^{6n_b}$ is the $6n_b$-dimensional space of stacked spatial force vectors~\cite{featherstone2014rigid}.
If the tree joints in \aggregate~joint $k$ permit a total of $n_f$ degrees of freedom and the loop constraints eliminate $n_l$ of those, then $\gS_k$ is a $6n_b \times (n_f-n_l)$ matrix, and the total number of constraints imposed by the \aggregate~joint is $n_c = 6n_b - (n_f-n_l)$.

The total force transmitted across \aggregateArticle~\aggregate~joint, $\gfJ$, describes forces applied on each body in \aggregateArticle~\aggregate~by its output body.
This net effect results from the constraint and active forces transmitted by the joints in $\calJ$.
Recall that this includes both spanning tree joints and loop joints.
The \aggregate~joint force can be expressed
\begin{equation}
    \gfJ_k  \triangleq \vcat{\agglink{k}}{\fJ_i} \triangleq \gforceSSM^a_k \gtauInd_k + \gforceSSM^c_k \gCnstrForceInd_k, \label{eqn:aggregate_joint_force_decomp}
\end{equation}
where $\gforceSSM^a_k \gtauInd_k$ is the active force and $\gforceSSM^c_k \gCnstrForceInd_k$ is the constraint force.
The active force is the product of the \aggregate~active force subspace matrix $\gforceSSM^a_k\in\R^{6n_b\times(n_f-n_l)}$ and the generalized active forces $\gtauInd_k$. 

Similarly, the constraint force is the product of the \aggregate~constraint force subspace matrix $\gforceSSM^c_k\in\R^{6n\times n_c}$ and the generalized constraint forces arising from the tree and loops joints $\gCnstrForceInd_k$.
The constraint force must lie in $\motionSS_k^\perp$, so $\gforceSSM^c_k$ must span the force subspace $\forceSS^c_k=\motionSS_k^\perp$.

Because the total force transmitted across \aggregateArticle~\aggregate~joint can be any spatial force in $\forceSpace^{6n}$, $\forceSS^a_k$ must satisfy $\forceSS^c_k\oplus\forceSS^a_k=\forceSpace^{6n}$.
Therefore, $\gforceSSM^a_k$ and $\gforceSSM^c_k$ must satisfy,
\begin{equation}
    \gS_k^\top \gforceSSM^a_k = \mathbf{1}, \quad \gS_k^\top \gforceSSM^c_k = \mathbf{0}, \label{eqn:aggregated_dual_basis}
\end{equation}
which, via~\eqref{eqn:aggregate_joint_force_decomp}, leads to the crucial relationship
\begin{equation}
    \gS_k^\top \gfJ_k = \gtauInd_k. \label{eqn:aggregate_active_force_from_transmitted}
\end{equation}
Similarly to Sec.~\ref{ssec:agg_motion_ss}, these \aggregate~force subspace matrices can be constructed from conventional force subspace matrices and constraint Jacobians.
To do so, we will again relate the quantities transmitted across the \aggregate~joint to quantities transmitted across individual tree joints.
A visualization of \aggregateArticle~\aggregate~joint transmitting forces $\fJ_i$ to the bodies in its child \aggregateLink~is shown in Fig.~\ref{fig:statically_equivalent_forces}a and~\ref{fig:statically_equivalent_forces}b.

First, consider \aggregateArticle~\aggregateLink~$k$. 
The net force on body $i$ in \aggregateLink~$k$ is equal to the force transmitted to it by the \aggregate~joint minus the forces it transmits across tree joints to its child bodies not in \aggregateLink~$k$.
In Fig~\ref{fig:statically_equivalent_forces}b, these forces are shown in green and gray, respectively.
The net force $\gfnet_k\triangleq\vcat{\agglink{k}}{\fnet_i}$ on the \aggregateLink~can therefore be written
\newcommand{\outOfClusterForces}{\sum_{j\in\childT{i}\setminus\agglink{k}}\hspace{-10pt}{}^i\fj_j}
\begin{equation}
    \gfnet_k = \vcat{\agglink{k}}{\fJ_i - \outOfClusterForces} \label{eqn:fJ_representation}
\end{equation}

The net force on body $i$ in \aggregateLink~$k$ can alternatively be expressed exclusively in terms of forces transmitted across individual tree joints.
In this representation, the net force on body $i$ is equal to the force transmitted across its parent tree joint minus the forces it transmits across tree joints to \textit{all} of its child bodies in $\spantree$.
In Fig~\ref{fig:statically_equivalent_forces}c, which shows a statically equivalent system to Fig~\ref{fig:statically_equivalent_forces}b, the forces between bodies in \aggregateLink~$k$ are shown in blue and forces transmitted to bodies not in \aggregateLink~$k$ are shown in gray.
The net force on the \aggregateLink~$k$ can, therefore, be written
\begin{equation}
    \gfnet_k = \vcat{\agglink{k}}{\fj_i - \sum_{j\in\childT{i}\cap\agglink{k}}\hspace{-10pt}{}^i\fj_j -\outOfClusterForces}. \label{eqn:fj_representation}
\end{equation}

To demonstrate that these representations are equivalent, we will find the matrix $\gSPOF{k}$ that relates the forces transmitted across individual tree joints in the spanning tree to forces transmitted across the \aggregate~joint.
The matrix is comprised of spatial force transforms between individual rigid bodies and satisfies the relationship,
\begin{equation}
    \vcat{\agglink{k}}{\fJ_i} = \gSPOF{k}\vcat{\agglink{k}}{\fj_i}. \label{eqn:spof}
\end{equation}

To find this matrix, we first note that the contribution of forces transmitted to bodies not in \aggregateLink~$k$ (gray forces in Fig.~\ref{fig:statically_equivalent_forces}) is equivalent for \eqref{eqn:fJ_representation} and~\eqref{eqn:fj_representation}.
Therefore, we have that
\begin{equation}
    \vcat{\agglink{k}}{\fJ_i} = \vcat{\agglink{k}}{\fj_i - \sum_{j\in\childT{i}\cap\agglink{k}}\hspace{-10pt}{}^i\fj_j},
\end{equation}
which reveals that every 6-row block of $\gSPOF{k}$ corresponding to body $i$ contains identity on the diagonal and $-\XF{i}{j}$ in the 6-column block where $i$ is the parent of $j$.
Thus, the general formula for $\gSPOF{k}$ is given by
\begin{equation}
    \gSPOF{k} = \begin{bmatrix}
        \mathbf{1} & -\X^*_{1,2} & \hdots & -\X^*_{1,n} \\
        \mathbf{0} & \mathbf{1} & & \vdots \\
        \vdots & & \ddots & -\X^*_{n-1,n} \\
        \mathbf{0} & \hdots & \mathbf{0} & \mathbf{1}
    \end{bmatrix},
\end{equation}
with
\[
\X^*_{i,j} = \begin{cases}
      \XF{\agglink{k}(i)}{\agglink{k}(j)}, & \text{if}\ \agglink{k}(i) = \parentT{\agglink{k}(j)} \\
      \mathbf{0}, & \text{otherwise}
    \end{cases}.
\]
We note that $\gSPOF{k}$ can be decomposed into $\gSPOF{k} = \left(\mathbf{1} - \gSKO{k}\right)$, where $\gSKO{k}$ is the Spatial Kernel Operator for the spatial force transforms in the \aggregateLink~\cite{jain2012multibody_part2}.
We further note the relationship with $\gSPO{k}$,
\begin{equation}
    \gSPO{k}^\top = \left(\mathbf{1} - \gSKO{k}\right)^{-1} = \left(\gSPOF{k}\right)^{-1}. \label{eqn:spo_sko_inv}
\end{equation}

Proceeding with the construction, we start with the tree joint representation using conventional force subspace matrices for tree joints,
\begin{equation}
\begin{split}
    \gfj &\triangleq \vcat{\agglink{k}}{\fj_i} \\
    &= \vcat{\agglink{k}}{\forceSSM^a_i\ttau_i + \forceSSM^c_i\cnstrForce_i} \\ 
    &= \Dcat{\agglink{k}}{\forceSSM^a_i}\vcat{\agglink{k}}{\ttau_i} + \\
    &\hspace{0.4in}\Dcat{\agglink{k}}{\forceSSM^c_i}\vcat{\agglink{k}}{\cnstrForce_i}
\end{split} \label{eqn:stacked_tree_joint_force_decomp}
\end{equation}
where $\forceSSM^a_i$ is the active force subspace matrix for tree joint~$i$ in the spanning tree, $\ttau_i$ the generalized active force, $\forceSSM^c_i$ the constraint force subspace matrix, and $\cnstrForce_i$ the generalized constraint force~\cite{featherstone2014rigid}.
Observe the difference in generalized active and constraint forces in~\eqref{eqn:aggregate_joint_force_decomp} versus~\eqref{eqn:stacked_tree_joint_force_decomp}.
The former is written in terms of generalized active forces for only the independent tree joints $\gtauInd_k$ and generalized constraint forces that capture the effect of both tree and loop joints $\gCnstrForceInd_k$.
The latter, on the other hand, is expressed in terms of the generalized active forces for every tree joint in the spanning tree and the generalized constraint forces for every tree joint in the spanning tree.
Accordingly, we use a matrix $\gP_k$ to lift the independent active forces to a dynamically equivalent set of spanning tree active forces and the matrix $\gP^c_k$ to project the generalized constraint forces $\gCnstrForceInd_k$ onto the spanning tree joints,
\begin{equation}
\begin{split}
    \vcat{\agglink{k}}{\ttau_i} &= \gP_k\gtauInd_k, \\
    \vcat{\agglink{k}}{\cnstrForce_i} &= \gP^c_k\gCnstrForceInd_k. \label{eqn:force_projections}
\end{split}
\end{equation}
The matrix $\gP_k$ must satisfy $\gG_k^\top\gP_k=\mathbf{1}$, and so the Moore-Penrose pseudoinverse of $\gG_k^\top$ is a natural choice.
For $\gP^c_k$, any full-rank matrix satisfying $\gG_k^\top\gP^c_k=\mathbf{0}$ will suffice.

As a more advanced note, it is worth acknowledging that any right inverse of $\gG_k^\top$ would also suffice for $\gP_k$. The underlying reason for this flexibility is that any solution in the nullspace of $\gG_k^\top$ represents a purely self-opposing internal set of forces within the \aggregateLink. Such forces do not affect the dynamics and can be added to any of the columns of $\gP_k$ without affecting its validity.

Finally, we relate the \aggregate~joint force to the equivalent spanning tree joint forces via~\eqref{eqn:spof},
\begin{equation}
\begin{split}
    \gfJ_k &= \gSPOF{k}\vcat{\agglink{k}}{\fj_i} \\
    &\overset{\eqref{eqn:stacked_tree_joint_force_decomp}}{=} \gSPOF{k}\Dcat{\agglink{k}}{\forceSSM^a_i}\vcat{\agglink{k}}{\ttau_i} + \\
    &\hspace{0.4in}\gSPOF{k}\Dcat{\agglink{k}}{\forceSSM^c_i}\vcat{\agglink{k}}{\cnstrForce_i} \\
    &\overset{\eqref{eqn:force_projections}}{=} \gSPOF{k}\Dcat{\agglink{k}}{\forceSSM^a_i}\gP_k\gtauInd_k + \\
    &\hspace{0.4in}\gSPOF{k}\Dcat{\agglink{k}}{\forceSSM^c_i}\gP^c_k\gCnstrForceInd_k.
\end{split}
\end{equation}
and arrive at the constructions,
\begin{equation}
\begin{split}
    \gforceSSM^a_k &= \gSPOF{k}\Dcat{\agglink{k}}{\forceSSM^a_i}\gP_k, \\
    \gforceSSM^c_k &= \gSPOF{k}\Dcat{\agglink{k}}{\forceSSM^c_i}\gP^c_k.
\end{split} \label{eqn:aggregate_forceSSM_construction}
\end{equation}
In Appendix~\ref{app:validate}, we validate that these constructions of $\gS$, $\gforceSSM^a$, and $\gforceSSM^c$ in~\eqref{eqn:aggregate_motionSSM_construction} and~\eqref{eqn:aggregate_forceSSM_construction} satisfy~\eqref{eqn:aggregated_dual_basis}.

\subsection{\Aggregate~Spatial Transforms}

Lastly, we discuss generalizing the concept of spatial transforms between local frames of individual rigid bodies to spatial transforms between local frames of \aggregateLink s.\footnote{This definition is only valid for a parent and child in $\aggtree$. A general definition between non-adjacent nodes exists but is not required for subsequent derivations and is therefore omitted for brevity.}
For notational brevity, we will use $\ell\triangleq\parentA{k}$ for the remainder of the section. 
\newcommand{\kparent}{\ell}
Assume we have \aggregateArticle~\aggregated~spatial motion vector $\gv_\kparent$ that describes the motion of the parent of \aggregateLink~$k$,
\[
\gv_\kparent = \vcat{\agglink{\kparent}}{\v_i}.
\]
When applied to \aggregateLink s, transforming \aggregateArticle~\aggregated~spatial motion vector involves transforming $\v_{\output{k}}$ to the local frame of every body in \aggregateLink~$k$.
We therefore define the transformed motion vector ${}^k\gv_\kparent$ by
\[
{}^k\gv_\kparent \triangleq \vcat{\agglink{k}}{^i\v_{\output{k}}},
\]
where body $\output{k}$ will always be in \aggregateLink~$\kparent$.

The \aggregate~spatial motion transform $\gXM{k}{\kparent}$ is the matrix satisfying ${}^k\gv_\kparent = \gXM{k}{\kparent}\gv_\kparent$.
This \aggregate~spatial motion transform serves the dual purpose of (i) performing coordinate transforms on the relevant quantities in \aggregateLink~$\kparent$ and (ii) accounting for the output body connectivity between the \aggregateLink s~$k$ and~$\kparent$.
We define the following indicator function $\EpsilonA{k}{\cdot}$ to describe connectivity between \aggregateLink s $k$ and $\kparent$, 
\begin{equation}
\EpsilonA{k}{i} = 
    \begin{cases}
      1, & \text{if}\ \output{k} = \agglink{\kparent}(i) \\
      0, & \text{otherwise}
    \end{cases},
\end{equation}
leading to the general formula,
\begin{equation}
\gXM{k}{\kparent} = \begin{bmatrix}
    \EpsilonA{k}{1} \XM{\agglink{k}(1)}{\agglink{\kparent}(1)} & \cdots & \EpsilonA{k}{m} \XM{\agglink{k}(1)}{\agglink{\kparent}(m)} \\
    \vdots & \ddots & \vdots \\
    \EpsilonA{k}{1} \XM{\agglink{k}(n)}{\agglink{\kparent}(1)} & \cdots & \EpsilonA{k}{m} \XM{\agglink{k}(n)}{\agglink{\kparent}(m)}
\end{bmatrix}. \label{eqn:cluster_spatial_transform}
\end{equation}
In the original constraint embedding derivation~\cite{jain2009recursive}, this matrix is split into two parts corresponding to the two purposes noted above: $\gXM{k}{\kparent} = \gSPO{k}{}\E_k$, where $\gSPO{k}$ is the previously defined Spatial Propagation Operator and $\E_k$ is a ``connector block" encoding the connectivity between \aggregateLink~$k$ and its output body.

In the case of \aggregated~spatial force vectors, for
\[
\gf_k = \vcat{\agglink{k}}{\f_i},
\]
transforming this vector involves collecting all of the forces in \aggregateLink~$k$, transforming them to the local frame of body $\output{k}$, and then applying their sum to body $\output{k}$.
We therefore define the transformed force vector ${}^\kparent\gf_k$ by
\[
{}^\kparent\gf_k \triangleq \vcat{\agglink{\kparent}}{\EpsilonA{k}{i}\sum_{j\in\agglink{k}}{}^{i}\f_j}.
\]
The matrix $\gXF{\kparent}{k}$ satisfying ${}^\kparent\gf_k=\gXF{\kparent}{k}\hspace{1pt}\gf_k$
has the general formula
\begin{equation}
\gXF{\kparent}{k} =\begin{bmatrix}
    \EpsilonA{k}{1}\XF{\agglink{\kparent}(1)}{\agglink{k}(1)} & \cdots & \EpsilonA{k}{1}\XF{\agglink{\kparent}(1)}{\agglink{k}(n)} \\
    \vdots & \ddots & \vdots \\
    \EpsilonA{k}{m} \XF{\agglink{\kparent}(m)}{\agglink{k}(1)} & \cdots & \EpsilonA{k}{m} \XF{\agglink{\kparent}(m)}{\agglink{k}(n)}
\end{bmatrix}.
\end{equation}

We emphasize two important properties of spatial transforms between \aggregateLink s.
First, they preserve the following crucial property of conventional spatial transforms: $\gXM{k}{\kparent}^\top=\gXF{\kparent}{k}$.
Intuitively, this property can be understood by the dual relationship between stacking motion vectors such that they can propagate outward to their subtrees and collecting force vectors such that they can propagate inward as net forces on shared output bodies.
Second, because $\gXF{\kparent}{k}$ has only one non-zero block of columns and $\gXM{k}{\kparent}$ has only one non-zero block of rows, left and right multiplying a matrix by \aggregateArticle~\aggregate~force transform and its dual \aggregate~motion transform, respectively, produces a block diagonal matrix.
More specifically, the only non-zero elements of the product will be the block diagonal elements corresponding to the output body of cluster $k$, as indicated by $\EpsilonA{k}{\cdot}$.

To demonstrate the outward propagation of motion vectors, consider a rearranged version of~\eqref{eqn:define_vJ}, where the stacked velocity vector for \aggregateLink~$k$ can be expressed in terms of its output body velocity and its velocity across the \aggregate~joint,
\begin{equation}
    \vcat{\agglink{k}}{\v_i} = \vcat{\agglink{k}}{{}^i\v_{\output{k}}} + \v^\calJ_k.
\end{equation}
This relationship can be concisely represented with the \aggregate~spatial transform defined in~\eqref{eqn:cluster_spatial_transform} and the \aggregate~motion subspace matrix defined in~\eqref{eqn:vagg_J}, 
\begin{equation}
\begin{split}
    \gv_k = \gXM{k}{\parentA{k}}\gv_\parentA{k} + \gS_k\gyd_k. \label{eqn:agg_velocity_prop}
\end{split}
\end{equation}
Likewise for the acceleration defined in~\eqref{eqn:agg_aJ},
\begin{equation}
    \ga_k = \gXM{k}{\parentA{k}}\ga_\parentA{k} + \gS_k\gydd_k + \gSd_k\gyd_k, \label{eqn:agg_accel_prop}
\end{equation}
which recovers common formulas from the classical open-chain case \cite{featherstone2014rigid}, but now with applicability in a broader setting.

%% file: Sections/PropagationMethodForRecursiveAlgorithmsUsingConstraintEmbedding.tex
\section{Propagation Perspective on Recursive Algorithms Using Local Constraint Embedding} \label{sec:cluster-propagation}

In this section, we provide a self-contained derivation of the recursive forward dynamics algorithm using local constraint embedding.
Central to our derivation is our method of developing articulated inertia for \aggregateLink s.
We develop this concept through connections to Featherstone's multi-handle assembly process~\cite{featherstone1999divide2}.
The resulting algorithm is ultimately identical to Jain's original recursive forward dynamics algorithm using local constraint embedding, but the derivation is distinct from the Riccati-Equation-based derivation developed in \gls{soa}~\cite{jain2009recursive,jain2012multibody_part2}.
Specifically, our derivation is the first assembly process to demonstrate the recursivity of articulated inertia computation in the presence of loop constraints.

\subsection{Articulated-Body Inertia for \AggregateLink s}

A rigid body's articulated-body inertia describes the inertia that the rigid body \textit{appears to have} when it is part of a rigid-body subsystem~\cite{featherstone1983calculation}.
In isolation, under the effect of force $\f_i$, body $i$ follows the spatial equation of motion
\begin{equation}
    \f_i = \I_i\mathbf{a}_i + \p_i,
\end{equation}
where $\p_i$ is a bias force. 
By contrast, if body $i$ is attached to a child subsystem via joints to form an ``articulated body," it will follow the spatial equation of motion
\begin{equation}
    \f_i = \IA_i\mathbf{a}_i + \pA_i, \label{eqn:articulated_body_inertia}
\end{equation}
where $\IA_i$ and $\pA_i$ are the multi-body system's articulated-body inertia and bias force, respectively.
Body $i$ is referred to as the ``handle" of the articulated body.
The handle is the body to which $\f_i$ is applied and to which $\IA_i$ and $\pA_i$ refer.

To use~\eqref{eqn:articulated_body_inertia}, the handle must have six degrees of freedom.
If it does not, its spatial equation of motion must instead be expressed in terms of its articulated-body \textit{inverse} inertia $\invIA_i$ and bias \textit{acceleration} $\bA_i$,
\begin{equation}
    \mathbf{a}_i = \invIA_i\f_i + \bA_i.
\end{equation}
An articulated body can have multiple handles.
For an articulated body with $h$ handles, the multi-handle spatial equations of motion nominally have the form~\cite{featherstone1999divide2}:
\begin{equation}
    \begin{bmatrix} \mathbf{a}_1 \\ \vdots \\ \mathbf{a}_h \end{bmatrix} = 
    \begin{bmatrix} \invIA_1 & \hdots & \invIA_{1,h} \\ \vdots & \ddots & \vdots \\ \invIA_{h,1} & \hdots & \invIA_h \end{bmatrix}
    \begin{bmatrix} \f_1 \\ \vdots \\ \f_h \end{bmatrix}
    + \begin{bmatrix} \bA_1 \\ \vdots \\ \bA_h \end{bmatrix}
    \label{eqn:multi_handle_invIA}
\end{equation}
where $\invIA_{i,j}$ is the cross-coupling inverse inertia between handles $i$ and $j$.
The combined $\invIA$ matrix is symmetric and positive definite, and its rank equals the total number of independent degrees of freedom possessed by the handles.
If the rank $6h$, then~\eqref{eqn:multi_handle_invIA} can be inverted to obtain a multi-handle generalization of~\eqref{eqn:articulated_body_inertia},
\begin{equation}
    \begin{bmatrix} \f_1 \\ \vdots \\ \f_h \end{bmatrix} = 
    \begin{bmatrix} \IA_1 & \hdots & \IA_{1,h} \\ \vdots & \ddots & \vdots \\ \IA_{h,1} & \hdots & \IA_h \end{bmatrix}
    \begin{bmatrix} \mathbf{a}_1 \\ \vdots \\ \mathbf{a}_h \end{bmatrix}
    + \begin{bmatrix} \pA_1 \\ \vdots \\ \pA_h \end{bmatrix}.
    \label{eqn:multi_handle_IA}
\end{equation}
Furthermore, if $\invIA_i$ is invertible for all handles and $\invIA_{i,j}=\mathbf{0}$ for all pairs of handles,
\begin{equation}
    \begin{bmatrix} \f_1 \\ \vdots \\ \f_h \end{bmatrix} = 
    \begin{bmatrix} \IA_1 & \hdots & \mathbf{0} \\ \vdots & \ddots & \vdots \\ \mathbf{0} & \hdots & \IA_h \end{bmatrix}
    \begin{bmatrix} \mathbf{a}_1 \\ \vdots \\ \mathbf{a}_h \end{bmatrix}
    + \begin{bmatrix} \pA_1 \\ \vdots \\ \pA_h \end{bmatrix}.
    \label{eqn:multi_handle_bdiag_IA}
\end{equation}
This block diagonal form corresponds to the case where every handle in the articulated body is the root body for its own distinct floating-base subsystem and thus has six degrees of freedom.

Now consider the case where an articulated body contains all bodies supported by the bodies in \aggregateLink~$k$.
Let the bodies in \aggregateLink~$k$ be the handles, and let any constraints (i.e., tree and loop joints) between those handle bodies be removed.
For this articulated body, the multi-handle equations of motion~\eqref{eqn:multi_handle_invIA} describe how the bodies in \aggregateLink~$k$ will accelerate with
\[
\ga_k \triangleq \vcat{\agglink{k}}{\mathbf{a}_i} \hspace{-2pt}=\hspace{-3pt} \begin{bmatrix} \mathbf{a}_1 \\ \vdots \\ \mathbf{a}_h \end{bmatrix}
\]
in response to applying forces
\[
\gf_k \triangleq \vcat{\agglink{k}}{\f_i} \hspace{-2pt}=\hspace{-3pt} \begin{bmatrix} \f_1 \\ \vdots \\ \f_h \end{bmatrix}
\]
to each of the bodies in \aggregateLink~$k$, considering the effects of the supported bodies.
Such an articulated body can always be expressed in the block-diagonal form~\eqref{eqn:multi_handle_bdiag_IA}, because each of the handles serves as the root for its own floating-base subsystem and therefore has six degrees of freedom.

\begin{figure}
    \centering
    \includegraphics[width=0.85\columnwidth]{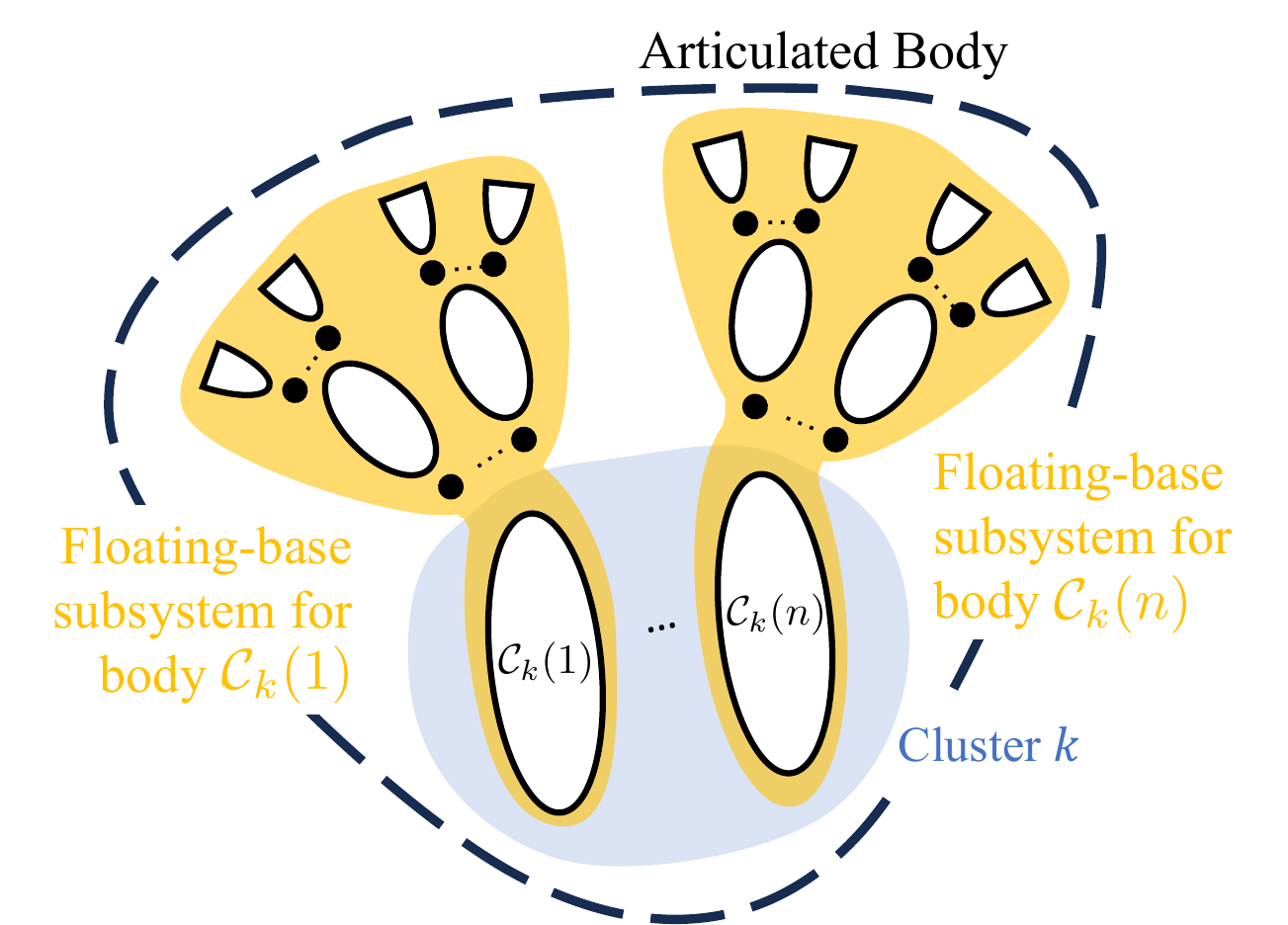}
    \caption{An articulated body where the multiple handles correspond to the bodies in \aggregateLink~$k$. Note that effect of constraints in $\calJ_k$ are not considered for this articulated body.}
    \label{fig:articulated_cluster}
\end{figure}

When the articulated body satisfies these criteria, the equations of motions for the handles (i.e., bodies in the \aggregateLink) can be written using \aggregate~notation
\begin{equation}
    \gf_k = \gIA_k\ga_k + \gpA_k. \label{eqn:articulated_agg_link_inertia}
\end{equation}
We show such an articulated body with block-diagonal $\gIA$ in Fig.~\ref{fig:articulated_cluster}.
In Sec.~\ref{ssec:agg_link_assembly}, we will show that assembling articulated bodies using \aggregate~joints is guaranteed to preserve the block-diagonality and well-posedness of $\gIA$, and in Sec.~\ref{ssec:physical_interpretation} we will further interpret the physical meaning of $\gIA_k$.

\begin{figure}
    \centering
    \includegraphics[width=\columnwidth]{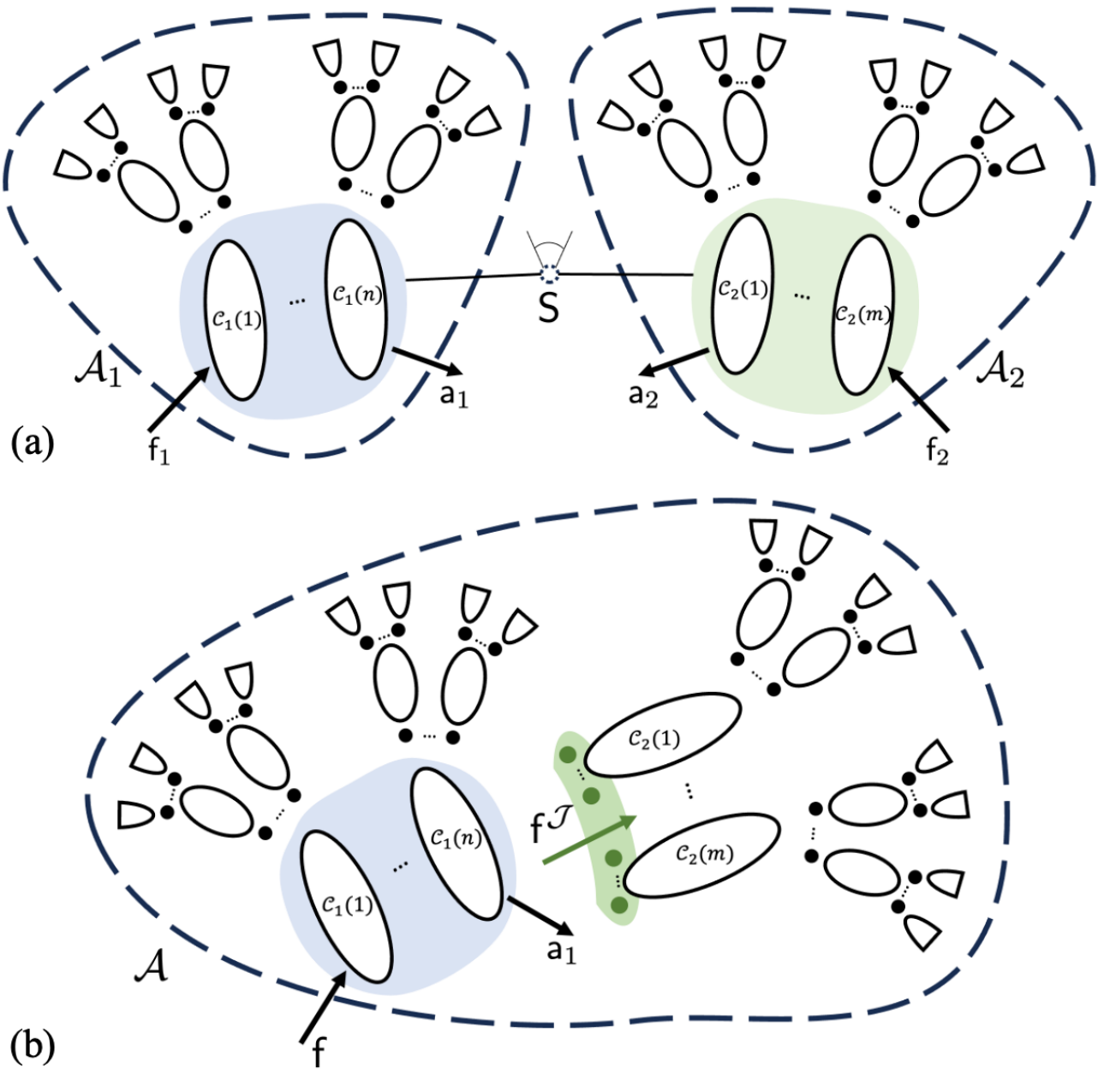}
    \caption{Assembly process for two articulated bodies having \aggregateLink~1 and \aggregateLink~2 as their handles, respectively. \textbf{(a)} Articulated bodies prior to assembly. \textbf{(b)} Articulated body assembled by joining the handles in (a) with \aggregateArticle~\aggregate~joint (shown in green). The \aggregate~joint captures the effects of any joints between bodies in \aggregateLink~1 and \aggregateLink~2, as well as any joints between bodies \textit{within} \aggregateLink~2.}
    \label{fig:assembling_articulated_bodies}
\end{figure}

\subsection{Assembling Articulated Bodies With \Aggregate~Joints} \label{ssec:agg_link_assembly}

Consider the two articulated bodies shown in Fig.~\ref{fig:assembling_articulated_bodies}a, $\AB_{1}$ and $\AB_{2}$.
They contain all of the bodies supported by the bodies in \aggregateLink s $1$ and $2$, respectively.
They have $n$ and $m$ handles, respectively, and these handles correspond to the bodies in \aggregateLink s $1$ and $2$.
The handles are the root bodies of distinct floating-base subsystems because the \aggregate~joints $\calJ_1$ and $\calJ_2$ have been removed.
We can, therefore, write their equations of motion,
\begin{equation}
\begin{split}
    \gf_1 &= \gIA_1\ga_1 + \gpA_1, \\
    \gf_2 &= \gIA_2\ga_2 + \gpA_2,  \label{eqn:ab1_2_eom}
\end{split}
\end{equation}
where $\gIA_1\in\R^{6n\times 6n}$ and $\gIA_2\in\R^{6m\times 6m}$ are the positive definite, block-diagonal articulated inertias for the \aggregateLink s.

We will next follow similar assembly steps to~\cite{featherstone1983calculation} but do so in a way that leverages constraint embedding.
The new articulated body, $\AB$, that results from assembling $\AB_{1}$ and $\AB_{2}$ is shown in Fig.~\ref{fig:assembling_articulated_bodies}b.
The assembly is done by connecting the handles of $\AB_{1}$ to the handles of $\AB_{2}$ through \aggregateArticle~\aggregate~joint.
The use of \aggregateArticle~\aggregate~joint to join articulated bodies distinguishes our assembly method from Featherstone's original multi-handle assembly method~\cite{featherstone1999divide2}.
Specifically, as we will demonstrate, using the \aggregate~joint guarantees the invertibility of~\eqref{eqn:multi_handle_invIA} into the form of \eqref{eqn:articulated_agg_link_inertia}.

Note that in Fig.~\ref{fig:assembling_articulated_bodies}b the handles of $\AB$ are still the bodies \aggregateLink~$1$, and they have six degrees of freedom because they are still the roots of distinct floating-base subsystems.
Thus, we can write the equations of motion for the handles of $\AB$ as
\begin{equation}
    \gf = \gIA\ga_1 + \gpA, \label{eqn:assembled_articulated_body_eom}
\end{equation}
where $\gIA\in\R^{6n\times 6n}$ is the articulated inertia of \aggregateLink~$1$ when its bodies act as the handles of $\AB$.
In a similar manner to~\cite{featherstone1983calculation}, we next show how to obtain $\gIA$ and $\gpA$ in terms of the known articulated quantities from~\eqref{eqn:ab1_2_eom}.

Recall that the force transmitted across \aggregateArticle~\aggregate~joint, $\gfJ$, represents the force on each body in its child \aggregateLink~emerging from active and constraint forces. 
Furthermore, these forces can be propagated back to output bodies in the parent \aggregateLink~of the cluster joint via \aggregateArticle~\aggregate~spatial force transform.
Thus, the handle forces on the pre-assembly articulated bodies can be related to the handle forces on the assembled articulated body by
\begin{equation}
\begin{split}
    \gf_1 &= \gf - \gXF{1}{2}\gfJ, \\
    \gf_2 &= \gfJ. \label{eqn:assembly_force_balance}
\end{split}
\end{equation}

The \aggregate~joint force imposes the motion constraints of both the tree and loop joints via~\eqref{eqn:agg_aJ},
\begin{equation}
    \gaJ = \ga_2 - \gXM{2}{1} \ga_1 = \gS\gydd + \gSd\gyd, \label{eqn:agg_assembly_acc_constraint}
\end{equation}
where $\gS$ is the known \aggregate~motion subspace matrix for the connecting \aggregate~joint and $\gydd$ is the unknown independent joint acceleration across the \aggregate~joint.
Furthermore, we know from~\eqref{eqn:aggregate_active_force_from_transmitted} that
\begin{equation}
    \gS^\top\gfJ = \gtauInd, \label{eqn:assembly_power_balance}
\end{equation}
where $\gtauInd$ is the generalized active force applied to the independent tree joints of the \aggregate~joint. 

The rest of the process mirrors the original propagation methods of the ABA in this more general context.
The unknown $\gydd$ can be solved for by substituting~\eqref{eqn:assembly_force_balance} into~\eqref{eqn:assembly_power_balance},
\begin{equation}
\begin{split}
    \gtauInd &= \gS^\top\gf_2 \\
    &\overset{\eqref{eqn:ab1_2_eom}}{=} \gS^\top\left(\gIA_2\ga_2 + \gpA_2 \right) \\
    &\overset{\eqref{eqn:agg_assembly_acc_constraint}}{=} \gS^\top\left(\gIA_2\left(\gXM{2}{1}\ga_1 + \gS\gydd + \gSd\gyd\right) + \gpA_2 \right)
\end{split}
\end{equation}
and isolating it such that
\begin{equation}
    \gydd = (\gS^\top\gIA_2\gS)^{-1}\left(\gtauInd - \gS^\top\left(\gIA_2\left(\gXM{2}{1}\ga_1 + \gSd\gyd\right) + \gpA_2 \right)\right) \label{eqn:assembled_ydd}
\end{equation}
Since $\gIA_2$ is positive definite and $\gS^\top$ is full column rank if the \aggregateLink~is not overconstrained, $(\gS^\top\gIA_2\gS)$ will always be invertible.

Lastly, we relate force acting on the handles of the $\AB$ to the acceleration of \aggregateLink~$1$,
\begin{equation}
\begin{split}
    \gf &\overset{\eqref{eqn:assembly_force_balance}}{=} \gf_1 + \gXF{1}{2}\gf_2 \\
    &\overset{\eqref{eqn:ab1_2_eom}}{=} \gIA_1\ga_1 + \gXF{1}{2}\gIA_2\ga_2 + \gpA_1 + \gXF{1}{2}\gpA_2 \\
    &\overset{\eqref{eqn:agg_assembly_acc_constraint}}{=} \gIA_1\ga_1 + \gXF{1}{2}\gIA_2\left(\gXM{2}{1}\ga_1 + \gSd\gyd + \gSd\gydd\right) + \gpA_1 + \gXF{1}{2}\gpA_2 \\
    &\overset{\eqref{eqn:assembled_ydd}}{=} \gIA_1\ga_1 + \gXF{1}{2}\gIA_2\left(\gXM{2}{1}\ga_1 + \gSd\gyd\right) + \gpA_1 + \gXF{1}{2}\gpA_2 + \\
    &\hspace{0.4in}\gXF{1}{2}\gIA_2\gS(\gS^\top\gIA_2\gS)^{-1}\gtauInd - \\
    &\hspace{0.4in}\gXF{1}{2}\gIA_2\gS(\gS^\top\gIA_2\gS)^{-1}\gS^\top\gIA_2\gXM{2}{1}\ga_1 - \\
    &\hspace{0.4in}\gXF{1}{2}\gIA_2\gS(\gS^\top\gIA_2\gS)^{-1}\gS^\top\left(\gIA_2\gSd\gyd + \gpA_2 \right),
\end{split}
\end{equation}
and group terms to yield the coefficients of the equation of motion in~\eqref{eqn:assembled_articulated_body_eom},
\begin{equation}
    \gIA = \gIA_1 + \gXF{1}{2}\gIA_2\gXM{2}{1} - \gXF{1}{2}\gIA_2\gS\left(\gS^\top\gIA_2\gS\right)^{-1}\gS^\top\gIA_2\gXM{2}{1}. \label{eqn:aggregate_link_articulated_inertia_propagation}
\end{equation}
and 
\begin{equation}
\begin{split}
    \gpA &= \gpA_1 + \gXF{1}{2}\gpA_2 + \gXF{1}{2}\gIA_2\gSd\gyd + \\
    &\hspace{0.4in} \gXF{1}{2}\gIA_2\gS(\gS^\top\gIA_2\gS)^{-1}\left(\gtauInd - \gS^\top\left(\gIA_2\gSd\gyd + \gpA_2 \right)\right). \label{eqn:aggregate_link_articulated_bias_propagation}
\end{split}
\end{equation}
By construction of $\AB_1$ and $\AB_2$, $\gIA_1$ and $\gIA_2$ were block diagonal and positive definite.
Recalling that left and right multiplication by $\gXF{1}{2}$ and $\gXM{2}{1}$, respectively, preserves block-diagonality, $\gIA$ is also block diagonal.
Furthermore, because~\eqref{eqn:aggregate_link_articulated_inertia_propagation} can alternatively be expressed
\[
    \gIA = \gIA_1 + \gXF{1}{2}\left(\gL_2^\top\gIA_2\gL_2\right)\gXM{2}{1}
\]
where $\gL_k^\top \triangleq \mathbf{1} - \gIA_k\gS_k\left(\gS_k^\top\gIA_k\gS_k\right)^{-1}\gS_k^\top $ and is full column rank, $\gIA$ is also positive definite.
Thus, we have arrived at a set of equations for propagating articulated inertia and bias force that closely resembles the original propagation equations (38) and (39) in~\cite{featherstone1983calculation} and is mathematically identical to (50) in~\cite{jain2012multibody_part2}.

\subsection{Physical Interpretation} \label{ssec:physical_interpretation}
The \aggregate~joint assembly method for deriving~\eqref{eqn:aggregate_link_articulated_inertia_propagation} and~\eqref{eqn:aggregate_link_articulated_bias_propagation} lends itself to the straightforward physical interpretation of the articulated inertia for \aggregateArticle~\aggregateLink.
In Featherstone's general divide and conquer approach~\cite{featherstone1999divide2}, the articulated body $\tilde{\AB}$ that results from assembling $\tilde{\AB_{1}}$ and $\tilde{\AB_{2}}$ can have any number of handles, and those handles can be connected to each other via joints.
Consequently, the resulting collection of $h$ handles may have less than $6h$ degrees of fredom.
In that case, their equations of motion must be expressed in the inverse-inertia form of~\eqref{eqn:multi_handle_invIA}.
This form is compatible with the parallelizable Divide and Conquer algorithm for forward dynamics~\cite{featherstone1999divide2} but not for the serialized recursive algorithms of interest in this work.
The combined articulated inverse inertia for these handles $\tilde{\boldsymbol{\Phi}}^A$ will be symmetric.
Physically, the off-diagonal blocks represent how forces on one handle in $\tilde{\AB}$ influence the motion of other handles in $\tilde{\AB}$.

We distinguish the general Featherstone approach from the \aggregate-joint assembly in Sec.~\ref{ssec:agg_link_assembly}, where we have enforced that after assembling $\AB_{1}$ and $\AB_{2}$, the new assembly $\AB$ has the same handles as $\AB_1$ but they are not connected to one another and are therefore guaranteed to have six degrees of freedom.
Thus, the articulated inertia form of~\eqref{eqn:assembled_articulated_body_eom} can be used.
In this case, the articulated inertia for the \aggregateLink~$\gIA$ always has a block diagonal structure.
Physically, the block diagonal elements describe how the bodies in the \aggregateLink~will respond to forces, taking into account their connections to their respective floating-base subsystems.
Furthermore, the block diagonal structure means that a force on one handle does not influence the motion of the other handles.
Those interactions are not captured until the next assembly step, where $\AB$ is joined with the body or \aggregateLink~that precedes vi\aggregateArticle~\aggregateArticle~\aggregate~joint that captures these effects.

\subsection{Constraint Embedding Articulated-Body Algorithm}

The \gls{aba} consists of three passes: a first outward pass to calculate velocity and bias terms, an inward pass to calculate the articulated-body inertias and bias forces, and a second outward pass to calculate the accelerations.
Using the \aggregate~joint model from Sec.~\ref{sec:agg_joint_model} and the assembly process from Sec.~\ref{ssec:agg_link_assembly}, these same passes can be executed on $\aggtree$ to recursively compute the forward dynamics of a system with loop constraints.
We describe the steps below, remarking that they are essentially identical to~\cite{jain2009recursive}.
We also illustrate our specific implementation of the algorithm with pseudo-code in Alg.~\ref{alg:aca}.

\input{Sections/Helpers/ConstraintEmbeddingAbaPseudocode}

While the steps described below resemble the original \gls{aba}, the computational cost of the constraint embedding version of the algorithm is no longer linear in the number of degrees of freedom in $\spantree$.
Furthermore, the computational cost will depend on the formulation of constraint Jacobians and biases.
However, the explicit constraint formulation cost is generally hard to quantify, so we report only on the algorithm's complexity \textit{assuming the explicit constraint has been formulated}.
As originally reported in~\cite{jain2012multibody_part2}, the constraint-embedding \gls{aba} is, in the worst case, quadratic in the total degrees of freedom of the largest \aggregateLink~and cubic in the number of independent degrees of freedom in the largest \aggregateLink.
However, if an upper bound is placed on the size of \aggregateLink s (i.e., the loop joints are required to be ``local"), the algorithm remains linear in the overall system size.

\subsubsection{Outward Pass 1} Per~\eqref{eqn:agg_velocity_prop}, the spatial velocities of every body in \aggregateArticle~\aggregateLink~can be computed if the spatial velocities of the bodies in its parent \aggregateLink~and the state of its \aggregate~joint are known,
\begin{equation}
    \gv_k = \gXM{k}{\parentA{k}}\gv_\parentA{k} + \gS_k\gyd_k, \quad \left(\gv_0=\mathbf{0}\right). \label{eqn:agg_vel_recursion}
\end{equation}
This information can be used to compute the velocity-product acceleration $\gSd_k\gyd_k$ via~\eqref{eqn:aggregate_velocity_product_accel} and the bias force $\gp_k$ for each \aggregateLink,
\begin{equation}
\begin{split}
    \gp_k &= \vcat{\agglink{k}}{\v_i \crf \I_i \v_i} \\
    &= \Dcat{\agglink{k}}{\v_i \crf} \Dcat{\agglink{k}}{\I_i}\gv_k \\
    &= \gv_k \crf \gI_k\gv_k.
\end{split}
\end{equation}

\subsubsection{Inward Pass} The next step computes the articulated-body inertias and bias forces, $\gIA_k$ and $\gpA_k$, for every \aggregateLink~in $\aggtree$.
The quantities $\gIA_k$ and $\gpA_k$ correspond to the articulated body $\AB_{k}$.
The bodies in \aggregateLink~$k$ are the handles of $\AB_{k}$.

These articulated bodies are assembled starting from the leaves of $\aggtree$.
Specifically, for a leaf \aggregateLink~$k$, the articulated body $k$ has the bodies of \aggregateLink~$k$ as its handles. 
The handles are not connected to one another by joints, and they have no children, so
\[
    \gIA_k=\Dcat{\agglink{k}}{\I_i}, \quad \gpA_k=\vcat{\agglink{k}}{\p_i}.
\]
The spatial inertia of every individual body in \aggregateLink~$k$ is positive definite, so $\gIA_k$ is positive definite for all leaves.

The assembly progresses inward such that $\AB_{k}$ is assembled by connecting \aggregateLink~$k$ to all of its children in $\aggtree$ via \aggregate~joints $\calJ_\ell$ for all $\ell\in\childA{k}$ according to the process in Sec.~\ref{ssec:agg_link_assembly}.
Applying equations~\eqref{eqn:aggregate_link_articulated_inertia_propagation} and~\eqref{eqn:aggregate_link_articulated_bias_propagation} for each of these connections leads to the recursive formulae comprising the inward pass
\begin{equation}
\begin{split}
    \gIA_k &= \gI_k + \sum_{\ell\in\childA{k}} \gXF{k}{\ell}\gIa_\ell\gXM{\ell}{k}, \\
    \gpA_k &= \gp_k + \sum_{\ell\in\childA{k}} \gXF{k}{\ell}\gpa_\ell,
\end{split}
\end{equation}
where
\begin{equation}
\begin{split}
    \gIa_\ell &= \gIA_\ell - \gIA_\ell\gS_\ell\left(\gS^\top_\ell\gIA_\ell\gS_\ell\right)^{-1}\gS^\top_\ell\gIA_\ell, \\
    \gpa_\ell &= \gpA_\ell + \gIa_\ell\gSd_\ell\gyd_\ell +  \gIA_\ell\gS_\ell\left(\gS^\top_\ell\gIA_\ell\gS_\ell\right)^{-1}\left(\gtauInd_\ell - \gS^\top_\ell\gpA_\ell\right).
\end{split}
\end{equation}

\subsubsection{Outward Pass 2} Similar to~\eqref{eqn:agg_vel_recursion}, the spatial accelerations of every body in \aggregateArticle~\aggregateLink~can be computed if the spatial velocities of the bodies in the \aggregateLink, the spatial accelerations of the bodies in its parent \aggregateLink, and the acceleration of the \aggregate~joint are known,
\begin{equation}
    \ga_k = \gXM{k}{\parentA{k}}\ga_\parentA{k} + \gSd_k\gyd_k + \gS_k\gydd_k, \quad \left(\ga_0=-\mathbf{a}_g\right) \label{eqn:agg_acc_recursion}
\end{equation}
where the fixed base is assumed to be accelerating upward at the acceleration of gravity to avoid explicitly accounting for gravitational forces. 
Thus, the second outward pass propagates the accelerations outward from the fixed base using~\eqref{eqn:assembled_ydd} and~\eqref{eqn:agg_acc_recursion}.

\subsubsection{Algorithm Implementation Details} 

Our implementation of the constraint-embedding ABA takes as inputs the system model, the joint positions, the joint velocities, and the generalized active forces associated with the spanning tree joints $\ttau$.
The system model contains information including link geometry, link inertia, joint types, loop constraint functions, $\spantree$, and $\aggtree$.
We allow the position input to be provided as either spanning or independent, $\gq$ or $\gy$, and likewise for the velocity input, $\gqd$ or $\gyd$, and the generalized active forces, $\gtau$ or $\gtauInd$.
Lines 3 and 5 in Outward Pass \#1 perform the conversion from independent to spanning if $\gy$ or $\gyd$ are provided.
Similarly, the algorithm can output either the spanning tree joint accelerations $\gqdd$ or the independent joint accelerations $\gydd$, based on the user's preference.

In Lines 8 and 9 of Outward Pass \#1, we borrow notation from~\cite{featherstone2014rigid} to describe the information computed for each of the spanning tree joints.
Based on the tree joint type (e.g., revolute, prismatic, spherical) and configuration, \verb~jcalc()~ returns the spatial motion transform $\X_{J(i)}$ describing the relative pose of frame $i$ relative to its ``home" pose when $\q_i=\mathbf{0}$, the motion subspace matrix $\S_i$, and the apparent derivative of the motion subspace matrix $\Sring_i$.
The spatial motion transform $\X_{T(i)}$ is a fixed quantity giving the transform from $\parentT{i}$ to the home pose of frame $i$.

Finally, Lines 11-16 of Outward Pass \#1 offer opportunities to leverage sparsity patterns.
For instance, any matrix formed using $\mathrm{diag}(\cdot)$ exhibits significant sparsity and is thus dealt with block-by-block.
Furthermore, assume \aggregateLink~$k$ contains $n_b$ bodies.
The matrix $\gSPO{k}$, computed in Line 11, is block unit lower triangular and thus has $\frac{n_b(n_b+1)}{2}$ non-zero 6-block elements in the worst case (depth of \aggregateLink~$k$ equals $n_b$) but only $n_b$ non-zero 6-block elements in the best case (depth of \aggregateLink~$k$ equals 1).
The matrix $\gXM{k}{\parentA{k}}$, computed in Line 13, has only $n_b$ non-zero 6-block elements, corresponding to the columns associated with the output body of \aggregateLink~$k$.
Because the system's connectivity is known prior to run time, we identify the non-zero blocks in $\gSPO{k}$ and $\gXM{k}{\parentA{k}}$, and use recursive equations to carry out matrix multiplications with them when evaluating Lines 12, 14, 15, and 16.

%% file: Sections/Helpers/ConstraintEmbeddingAbaPseudocode.tex
\begin{algorithm}
  \caption{Constraint-Embedding Articulated-Body Algorithm (Our Implementation)}
  \label{alg:aca}
  \textbf{Inputs:} System model, $\gq$ \textit{OR} $\gy$, $\gqd$ \textit{OR} $\gyd$, $\gtau$ \textit{OR} $\gtauInd$ \\
  \textbf{Output:} $\gydd$ \textit{OR} $\gqdd$ \vspace{0.2cm} \\
  \textbf{Outward Pass \#1}
  \begin{algorithmic}[1]
      \State $\gv_0 = \mathbf{0}$
      \For{every node $k$ in $\aggtree$ (ascending order)}
        \State $\gq_k = \ggamma(\gy_k)$\hspace{0.5cm}\textit{(If needed)}
        \State $\gG_k$ = updateConstraintJacobian($k$, $\gq_k$)
        \State $\gqd_k = \gG_k\gyd_k$\hspace{0.5cm}\textit{(If needed)}
        \State $\gg_k$ = updateConstraintBias($k$, $\gq_k$, $\gqd_k$)
        \For{$i \in \agglink{k}$}
            \State $[\X_{J(i)}, \S_i, \Sring_i] =$ jcalc(jtype($i$), $\q_i$, $\qd_i$)
            \State $\XM{i}{\parentT{i}} = \X_{J(i)}\X_{T(i)}$
        \EndFor
        \State Compute $\gSPO{k}$ using $\XM{i}{\parentT{i}}\,\forall i\in\agglink{k}$ via~\eqref{eqn:internal_motion_transform}
        \State $\gS_k = \gSPO{k}\Dcat{\agglink{k}}{\S_i}\gG_k$
        \State Compute $\gXM{k}{\parentA{k}}$ using $\XM{i}{\parentT{i}}\,\forall i\in\agglink{k}$ via~\eqref{eqn:cluster_spatial_transform}
        \State $\gv_k = \gXM{k}{\parentA{k}}\gv_{\parentA{k}} + \gSPO{k}\Dcat{\agglink{k}}{\S_i}\gqd_k$
        \State $\gSd_k\gyd_k = \gSPO{k}\Dcat{\agglink{k}}{\S_i}\gg_k +$
        \State $\hspace{1.5cm}\gSPO{k}\Dcat{\agglink{k}}{\Sring_i + \v_i \crm \S_i}\gqd_k$
        \State $\gIA_k = \gI_k$
        \State $\gpA_k = \gv_k\crf\gI_k\gv_k$
      \EndFor
  \end{algorithmic}

  \vspace{0.1cm}\textbf{Inward Pass}
  \begin{algorithmic}[1]
      \For{every node $k$ in $\aggtree$ (descending order)}
        \State $\gtauInd_k = \gG_k^\top\gtau_k$ \hspace{0.5cm}\textit{(If needed)}
        \State $\gU_k = \gIA_k\gS_k$
        \State $\gD_k = \gS_k^\top\gU_k$
        \State $\gu_k = \gtauInd_k - \gS_k^\top\gpA_k$
        \If{$\parentA{k} \neq 0$}
            \State $\gIa = \gIA_k - \gU_k\gD_k^{-1}\gU_k^\top$
            \State $\gpa = \gpA_k + \gIa\gSd_k\gyd_k + \gU_k\gD_k^{-1}\gu_k$
            \State $\gIA_{\parentA{k}} = \gIA_{\parentA{k}} + \gXF{\parentA{k}}{k}\,\gIa\,\gXM{k}{\parentA{k}}$
            \State $\gpA_{\parentA{k}} = \gpA_{\parentA{k}} + \gXF{\parentA{k}}{k}\,\gpa$
        \EndIf
      \EndFor
  \end{algorithmic}
      
  \vspace{0.1cm}\textbf{Outward Pass \#2}
  \begin{algorithmic}[1]
      \State $\ga_0 = -\mathbf{a}_g$
      \For{every node $k$ in $\aggtree$ (ascending order)}
        \State $\ga' = \gXM{k}{\parentA{k}} \ga_\parentA{k} + \gS_k\gydd_k + \gSd_k\gyd_k$
        \State $\gydd_k = \gD_k^{-1}\left(\gu_k - \gU_k^\top\ga'\right)$
        \State $\ga_k = \ga' + \gS_k\gydd_k$
      \EndFor
  \end{algorithmic}

  \vspace{0.1cm}\textbf{Conversion to Spanning Tree Accelerations}\hspace{0.5cm}\textit{(If needed)} 
  \begin{algorithmic}[1]
      \For{every node $k$ in $\aggtree$}
        \State $\gqdd_k = \gG_k\gydd_k + \gg_k$
      \EndFor
  \end{algorithmic}
\end{algorithm}

%% file: Sections/RNEA.tex
\section{Connections to Recursive Newton Euler Algorithm} \label{sec:rnea_connection}

The RNEA is an efficient algorithm for computing the inverse dynamics of kinematic trees.
It consists of two passes: (1) an outward pass for computing the acceleration of each body in the tree and the corresponding force needed to produce it and (2) an inward pass to compute the forces transmitted across the joints needed to produce the net forces computed in the first pass.
The forces transmitted across each joint are used to compute the generalized active force required at each joint.

The system information required to carry out the RNEA is the system's connectivity, motion subspace matrices, and spatial inertias.
In Sec.~\ref{sec:agg_joint_model}, we developed generalizations of this information for \aggregate~trees and can thus apply them to yield a constraint-embedding version of the RNEA.
The algorithm is not new.
The development of constraint embedding in the \gls{soa} context guarantees that algorithms for kinematic trees transfer to \aggregate~trees~\cite{jain2012multibody_part2}.
Furthermore, Kumar et al.~\cite{kumar2020analytical} proposed a conceptually similar modular RNEA for dealing with series-parallel robots.
Instead, we include the algorithm as a demonstration of the ability of our \aggregate~joint model to be used in wider contexts and to emphasize how intermediate quantities in the algorithm take on a different physical meaning in the cluster context while still following the same algorithmic recurrences.

\subsection{Outward Pass}
Per~\eqref{eqn:agg_velocity_prop} and~\eqref{eqn:agg_accel_prop}, the spatial velocities and accelerations of every body in \aggregateArticle~\aggregateLink~can be computed if the spatial velocities and accelerations of the bodies in its parent \aggregateLink and the state of its \aggregate~joint are known,
\begin{equation}
\begin{split}
    \gv_k &= \gXM{k}{\parentA{k}}\gv_\parentA{k} + \gS_k\gyd_k, \quad (\gv_0 = \mathbf{0}), \\
    \ga_k &= \gXM{k}{\parentA{k}}\ga_\parentA{k} + \gSd_k\gyd_k + \gS_k\gydd_k , \quad (\ga_0 = -\mathbf{a}_g),
\end{split}
\end{equation}
where the fixed-based is again assumed to be accelerating upward at the acceleration of gravity to avoid explicitly accounting for gravitational forces.
Concatenating these accelerations and the velocity-product terms leads to the net force acting on the cluster
\begin{equation}
\begin{split}
    \gfnet_k &= \vcat{\agglink{k}}{\I_i\mathbf{a}_i + \v_i \crf \I_i\v_i} \\
    &= \gI_k\ga_k + \gv_k \crf \gI_k\gv_k.
\end{split}
\end{equation}

\subsection{Inward Pass}
The \aggregate~joint for \aggregateLink~$k$ transmits the \aggregate~spatial force $\gfJ_k$ to the bodies in \aggregateLink~$k$ (e.g., Fig.~\ref{fig:statically_equivalent_forces}).
In response, an equal and opposite force is generated on the parent \aggregateLink~of the \aggregate~joint.
Because \aggregate~joints transmit forces from an output body to all bodies in the child \aggregateLink, the equal and opposite force is experienced by only the output body.
This output body relationship for transmitting forces is captured by $\gXF{\parentA{k}}{k}$.

Therefore, the net force on \aggregateArticle~\aggregateLink~can be expressed
\begin{equation}
    \gfnet_k = \gfJ_k - \sum_{\ell\in\childA{k}} \gXF{k}{\ell}\gfJ_\ell,
\end{equation}
or, in recursive form,
\begin{equation}
    \gfJ_k = \gfnet_k + \sum_{\ell\in\childA{k}} \gXF{k}{\ell}\gfJ_\ell. \label{eqn:crnea}
\end{equation}
At the leaves of $\aggtree$, $\gfJ_k = \gfnet_k$, so~\eqref{eqn:crnea} can be propgated inward to compute $\gfJ$ for every \aggregateLink.
Finally, using the \aggregate~motion subspace matrix and~\eqref{eqn:aggregate_active_force_from_transmitted}, the generalized active force $\gtauInd$ for each \aggregateLink~can be computed via $\gtauInd_k = \gS_k^\top \gfJ_k$.
When $\gy_k$ corresponds to a set independent coordinates for the \aggregateLink, and those coordinates coincide with those of the actuated joints, then $\gtauInd_k$ gives the efforts of those actuators\footnote{In the more general case (e.g., when loop joints are actuated, \aggregateLink~coordinates do not coincide with actuator DoFs, or when the \aggregateLink~is overactuated), solutions for the actuator efforts $\bar{\uptau}_k$ are given by any solution to $\mathsf{W}_k^\top \bar{\uptau}_k = \gtauInd_k $ (see Footnote \ref{foot:general_actuation}).}.

%% file: Sections/Benchmark.tex
\section{Benchmarking} \label{sec:benchmark}

In this section, we will benchmark the constraint-embedding Articulated-Body Algorithm against a state-of-the-art non-recursive algorithm for computing constrained forward dynamics.
The goal of the benchmark is to demonstrate for which system topologies (e.g., number of bodies, constraint dimension, and constraint locality) the constraint-embedding algorithm excels. 
A similar analysis was carried out in~\cite{jain2012efficient}.
However, in this paper, we perform additional benchmarking that (1) applies the constrained forward dynamics algorithms to popular actuation sub-mechanisms in modern robotic systems and (2) compares against the ``Proximal and Sparse" algorithm for constrained forward dynamics~\cite{carpentier2021proximal}, which has been developed since the publication of~\cite{jain2012efficient}.

The Proximal and Sparse algorithm works by first formulating the constrained dynamics of a rigid-body system in minimal coordinates as a convex optimization problem using the Gauss Principle of Least Constraint~\cite{gauss1829neues}.
A proximal algorithm is used to solve this convex optimization problem, leading to a set of regularized Karush-Kuhn-Tucker conditions that guarantee the problem is well-posed.
The forward dynamics are computed by performing a sparsity-exploiting Cholesky factorization on the Karush-Kuhn-Tucker matrix.
The computational cost of the Proximal and Sparse algorithm, therefore, depends primarily on the factors that determine the size of the Karush-Kuhn-Tucker (number of bodies and constraint dimension) rather than on the locality of the constraints.

Our benchmarking code is available at~\cite{grbda2023}.
Like Pinocchio, our entire repository is compatible with the algorithmic differentiation framework CasADi~\cite{andersson2019casadi}.
For our benchmark, we use CasADi to symbolically evaluate the forward dynamics functions of interest, and then we compare the number of ``instructions" (i.e., basic arithmetic operations) required to evaluate the respective functions.
We recognize that comparing instruction count is different from comparing computation time and that a lower count does not necessarily mean a shorter computation time.
However, we believe that instruction count is a better metric for our benchmark because we are interested in comparing how the costs of the respective forward dynamics algorithms scale with system topology rather than comparing the performance of rigid-body dynamics libraries.
For a comparison of absolute computation time, we could implement techniques suggested by the Pinocchio developers~\cite{carpentier2019pinocchio} such as static polymorphism via Curiously Repeating Template Patterns and code generation.
This, however, is left as future work.
Finally, we note that even though our algorithm offers different options for inputs and outputs, we ensure that in all of our benchmarks, the algorithms being compared have the same inputs and outputs.
Specifically, we use spanning-tree positions, velocities, and active forces.

\input{Sections/Helpers/NumberOfClusterFigures}
\subsection{Cost With Respect to Number of \AggregateLink s}

Our first benchmarks analyze how the computational cost of the respective forward dynamics algorithms vary with the number of \aggregateLink s used to describe the rigid-body system.
We set up these benchmarks as follows.
First, we select the type of actuation sub-mechanism we will use to create the loop constraints on the system.
Next, we determine the depth $d_a$ and number of branches $b_a$ that the \aggregate~tree will have.
Finally, we construct the rigid-body system by chaining $d_a$ of the actuation sub-mechanisms in series for each of the $b_a$ branches.
This process is illustrated in Fig.~\ref{fig:number_of_cluster_setup}.
Thus, for a system whose actuation sub-mechanism leads to \aggregateLink s containing $n$ rigid bodies related by a constraint of dimension $m$, the system will contain $n\times d_a\times b_a$ individual rigid bodies and be subject to $m\times d_a\times b_a$ constraints.

The results of varying the sub-mechanism type, \aggregate~tree depth, and \aggregate~tree branching are shown in Fig.~\ref{fig:number_of_cluster_instr}.
We use three actuation sub-mechanisms for this benchmark: Link and Rotor, Parallel Belt Transmission, and Four-Bar Mechanisms.
The Link and Rotor sub-mechanism enforces a proportional relationship between the parent joints of a link and rotor pair (bodies $L$ and $R$ in Fig.~\ref{fig:number_of_cluster_setup}b) and thus results in \aggregateArticle~\aggregateLink~containing two rigid bodies.
The Parallel Belt Transmission is slightly more complicated, enforcing a standard Link and Rotor relationship between two bodies (bodies $L_1$ and $R_1$ in Fig.~\ref{fig:number_of_cluster_setup}b) as well as an additional proportional relationship between another rotor and a distal link (bodies $L_2$ and $R_2$ in Fig.~\ref{fig:number_of_cluster_setup}b).
The constraint between $L_1$ and $R_1$ is nested by the constraint between $L_2$ and $R_2$, resulting in \aggregateArticle~\aggregateLink~containing four rigid bodies.
The constraints of both of these sub-mechanisms are naturally represented as explicit constraints with constant Jacobians, $\gqdd = \gG\gydd$.
The Four-Bar mechanism relates the motion of an input and output link (bodies $I$ and $O$ in Fig.~\ref{fig:number_of_cluster_setup}b) by attaching a connecting rod (body $C$) to each link via revolute joints.
Unlike the other two mechanisms, this constraint is naturally represented as an implicit constraint with a state-dependent Jacobian, $\gK(\gq)\gqdd = \gk(\gq,\gqd) $.
Therefore, the explicit constraint matrices $\gG$ and $\gg$ must be extracted from $\gK$ and $\gk$ every time the constraint-embedding \gls{aba} is called.
The instructions required by this extraction are included in the count for the constraint-embedding \gls{aba}.
The Proximal and Sparse algorithm, however, works directly with $\gK$ and $\gk$.

The most important result of this benchmark is the validation that the constraint-embedding \gls{aba} has linear cost in the number of \aggregateLink s when an upper bound is placed on the number of bodies allowed in each \aggregateLink.
In this case, those upper bounds are 2, 4, and 3 for the Link and Rotor, Parallel Belt Transmission, and Four-Bar Mechanism, respectively.
The cost of the non-recursive Proximal and Sparse algorithm, on the other hand, grows at a rate greater than linear in the number of bodies and the constraint dimension, so as the depth of the \aggregate~tree increases, the benefit of using constraint embedding increases.

\input{Sections/Helpers/SizeOfClusterFigures}
\subsection{Cost With Respect to Locality of \AggregateLink s}

While the constraint-embedding \gls{aba} scales well when loop constraints are local, its efficiency diminishes as the size of the loops increases.
Considering this, our second benchmarks analyze the effect of increasing the loop size while keeping the number of rigid bodies and the constraint dimension constant.
We preface this benchmark by remarking that constraint embedding and the Proximal and Sparse approach are not at odds with one another. 
On the contrary, the ideal setup for simulating rigid-body systems would simultaneously use constraint embedding for local loops and Proximal and Sparse techniques for non-local loops.
Such an approach is possible but not implemented in this work.
Instead, we present a benchmark that is informative in terms of cost scaling but suboptimal in terms of practicality.

We set up these benchmarks as follows.
First, we construct a spanning tree with two branches, each having depth $d_t$.
Next, we add a loop constraint between the first two bodies in each branch, compute the constrained forward dynamics, and count the number of instructions required by the various algorithms.
We then move the loop constraint outward from the base to the next pair of bodies in each branch, and we again compute and count the number of instructions.

Two types of loop constraints are considered: transmission and connecting rod.
When the transmission constraint is added between the $d_l$-th body in each branch, it enforces a proportional relationship between the joint angles of the first $d_l$ joints in each branch,
\begin{equation}
    \sum_{i=1}^{d_l} \q_{1,i} = \eta \sum_{i=1}^{d_l} \q_{2,i}
\end{equation}
where $\q_{1,i}$ and $\q_{2,i}$ are the joint angles of the $i$th joint in the first and second branches, respectively, and $\eta$ is the proportional constant.
Thus, for a loop constraint at depth $d_l$, the loop size is $2d_l$.
The constraint is naturally represented as an explicit constraint and has a constant Jacobian.
An illustration of the transmission-constrained system is shown in Fig.~\ref{fig:size_of_cluster_setup}a.

The connecting rod constraint couples the motion of the branches by attaching a link to the $d_l$-th body in each branch via revolute joints.
Thus, for a loop constraint at depth $d_l$, the loop size is $2d_l+1$.
The constraint is most naturally represented as an implicit constraint, and numerical extraction of explicit constraint Jacobians is required for the constraint-embedding ABA.
An illustration of the connecting-rod-constrained system is shown in Fig.~\ref{fig:size_of_cluster_setup}b.

The results from repeating these benchmarks for $d_t=5,10,20,40$ are shown in Fig.~\ref{fig:size_of_cluster_instr}.
First, observe that the cost of the constraint-embedding \gls{aba} scales poorly with respect to $d_l$ (note the logarithmic scale of the y-axis).
This is because the number of bodies in the largest \aggregateLink~increases linearly with $d_l$, and the constraint-embedding \gls{aba} scales poorly as the sizes of the \aggregateLink s increase. 

For small values of $d_l$, the constraint-embedding \gls{aba} consistently requires fewer instructions than the Proximal and Sparse algorithm.
As $d_l$ increases, the Proximal and Sparse algorithm eventually beats it.
This occurs because, for fixed $d_t$ and fixed constraint dimension, the size of the Karush-Kuhn-Tucker matrix that needs to be factorized by the Proximal and Sparse algorithm remains fixed, and so the cost of the algorithm is less sensitive to changes in $d_l$.
This is especially true in the case of transmission-constrained systems because the implicit constraint Jacobian is state-independent.
However, note that the intersection points between the constraint-embedding \gls{aba} and the Proximal and Sparse algorithm occur at larger values of $d_l$ as $d_t$ increases.
This is because the increased depth of the spanning tree increases the size of the Karush-Kuhn-Tucker matrix, leading to a more expensive factorization.
For the constraint-embedding \gls{aba}, on the other hand, the recursivity of the algorithm is well suited to handle added depth.

\begin{remark}
While we do not include it in the benchmark due to the lack of an available implementation, the method of Brandl et al.~\cite{brandl1987algorithm} is a notable candidate to consider alongside constraint embedding and the proximal and sparse method in future algorithms that may seek to combine them.
Brandl's method likewise extends the original ABA but does so by adding recursive computations for operational space inverse inertias, extended force propagators~\cite{wensing2012reduced}, and acceleration biases associated with constraint directions at loop joints.
When loops are isolated, these propgations are also local, as they are only carried out only until reaching a common ancestor of the bodies in each loop joint.
In this respect, Brandl's algorithm remains linear in the number of bodies when loops are isolated regardless of the loop depth.
While it would thus be expected to be favorable for large $d_l$ in our previous example, its extra computation demands at each recursive step suggest that constraint embedding would likely still be preferable for common low-depth loops.    
\end{remark}

\subsection{Effects of Ignoring Actuator Dynamics}

A popular alternative approach to using constrained dynamics algorithms is to approximate or ignore these effects of actuation sub-mechanisms.
In the context of the graphical system modeling from Sec.~\ref{ssec:modeling}, this is equivalent to removing (not clustering) nodes and edges from the connectivity graph so that you are left with a kinematic tree. 
The inertial properties of remaining nodes may be modified to account for the removed bodies.
MuJoco, for example, allows adding a diagonal ``armature matrix" to the joint-space inertia matrix to approximate the effects of geared motors~\cite{todorov2012mujoco} while still using efficient recursive dynamics algorithms.
This is equivalent to removing all nodes and edges associated with rotors from the connectivity graph, and then adding their inertia, scaled by the gear ratio, to their respective links.
Techniques such as domain randomization~\cite{peng2018sim} can make policies trained with these approximate dynamics robust to such modeling errors, but typically at the cost of sample complexity and the optimality of the policy.

\begin{figure}
    \centering
    \includegraphics[width=\columnwidth]{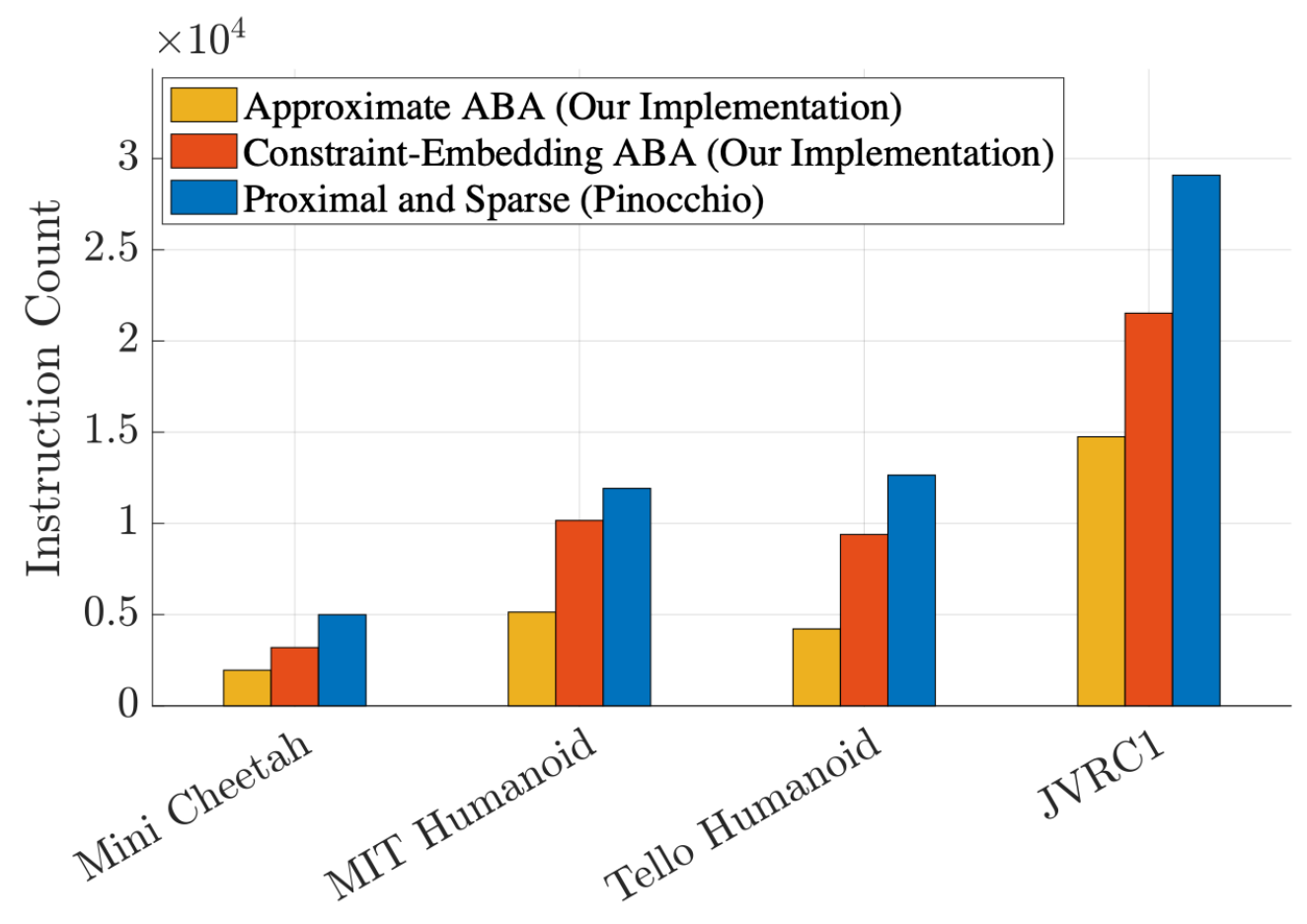}
    \caption{Relative cost of evaluating fully constrained vs. ``approximate" forward dynamics for a collection of legged robots.}
    \label{fig:approximate_instr}
\end{figure}

For our third benchmark, we compare the computational cost of the constraint-embedding \gls{aba} applied to a constrained system versus the standard \gls{aba} applied to the corresponding approximate system.
We perform this comparison for four legged robots using common actuation sub-mechanisms: the 12-DoF MIT Mini Cheetah~\cite{katz2019mini}, the 18-DoF MIT Humanoid~\cite{chignoli2021humanoid}, the 18-DoF Tello Humanoid~\cite{sim2022tello}, and a modified version of the 32-DoF JVRC-1 Humanoid~\cite{okugawa2015proposal}.
The MIT Mini Cheetah uses 12 link and rotor sub-mechanisms for its 12 actuated DoFs.
The MIT Humanoid uses 14 link and rotor sub-mechanisms for 14 of its actuated DoFs, and 2 parallel belt transmission sub-mechanisms for its remaining 4 actuated DoFs.
The Tello Humanoid uses 10 link and rotor sub-mechanisms for 10 of its actuated DoFs and 4 differential transmission sub-mechanisms for its remaining 8 actuated DoFs.
The JVRC-1 Humanoid is a generic highly actuated humanoid robot. We consider the version of the robot where 25 of its major links are actuated with rotor sub-mechanisms.

The results of the computational cost benchmark are shown in Fig.~\ref{fig:approximate_instr}.
For each of the systems analyzed, the Approximate \gls{aba} requires the fewest instructions: $38.8\%$ less than constrained-embedding \gls{aba} for the Mini Cheetah, $49.4\%$ less than constrained-embedding \gls{aba} for the MIT Humanoid, $55.1\%$ less than constrained-embedding \gls{aba} for the Tello Humanoid, and $31.5\%$ less than constrained-embedding \gls{aba} for the modified JVRC-1 Humanoid.
This reduction is expected because the approximate \gls{aba} deals with a simplified connectivity graph.
Furthermore, the instruction count reduction for the approximate \gls{aba} is more dramatic for the MIT Humanoid and Tello Humanoid than the Mini Cheetah and JVRC-1 because the Mini Cheetah and JVRC-1's \aggregateLink s contain fewer bodies.

However, the instruction count reduction for the approximate method comes at the cost of accuracy.
For our final benchmark, we analyze this trade-off with the following numerical experiment using one of the legs of the MIT Humanoid, shown in Fig.~\ref{fig:sub_mechanisms}.
Let us first distinguish between three types of systems: original, unconstrained, and approximate.
The original system contains all of the rigid bodies and constraints of the robot of interest.
The unconstrained system removes some of the rigid bodies and joints (rotors in the case of the MIT Humanoid Leg) so that the robot can be represented as a kinematic tree.
The approximate system is the same as the unconstrained but with modifications made to the remaining rigid bodies' inertia to account for the removed bodies (adding scaled rotor inertias to corresponding links in the case of the MIT Humanoid Leg.)

\begin{figure}
    \centering
    \includegraphics[width=0.9\columnwidth]{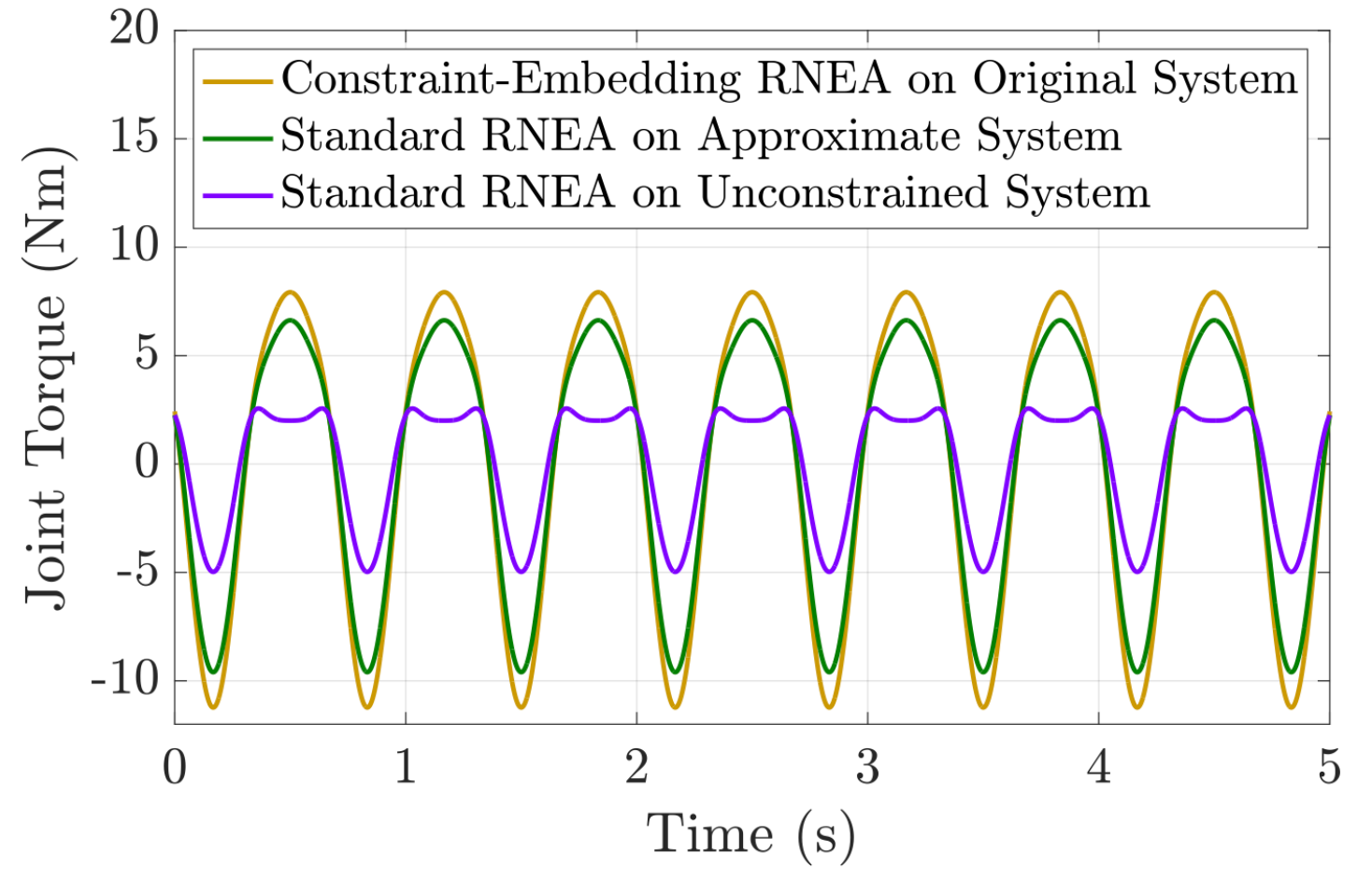}
    \caption{Comparison of inverse dynamics-based torque profiles required to achieve sinusoidal trajectories of the MIT Humanoid's ankle joint with $A=\SI{0.5}{\radian}, \omega=\SI{1.5}{\radian\per\second}$.}
    \label{fig:tracking_torque_comparison}
\end{figure}

To compare these approaches, we prescribe sinusoidal trajectories
\[
\begin{split}
\q(t) &= A\sin(2\pi\omega t), \\
\qd(t) &= 2A\pi\omega\cos(2\pi\omega t), \\
\qdd(t) &= -4A\pi^2\omega^2\sin(2\pi\omega t)
\end{split}
\]
for each of the joints of the leg.
Every $\Delta t$ seconds, we use the following algorithms to compute the torques required to induce the desired acceleration: constraint-embedding \gls{rnea} applied to the original system, standard \gls{rnea} applied to the unconstrained system, and standard \gls{rnea} applied to the approximate system.
The results in Fig.~\ref{fig:tracking_torque_comparison} show the difference in the required torque computed by the various algorithms. 

This numerical experiment validates that neglecting or even approximating the effects of loop constraints arising from actuation sub-mechanisms can lead to significant errors.
For example, the results in Fig.~\ref{fig:tracking_torque_comparison} show that using standard \gls{rnea} on the unconstrained system to produce the feedforward torque trajectory for the ankle joint yields a root mean square of \SI{4.31}{\newton\meter} relative to constraint-embedding \gls{rnea} on the original system.
Standard \gls{rnea} on the approximate system, which requires only a negligible amount of additional computation, reduces that root mean square error to \SI{1.04}{\newton\meter}.
These errors are $39.2\%$ and $9.5\%$ of the maximum torque, respectively. 
We chose to focus on the ankle joint in this experiment because its couplings with the ankle rotor joint, knee joint, and knee rotor joint are especially difficult for approximate algorithms to account for.

\begin{figure*}
    \centering
    \includegraphics[width=500pt]{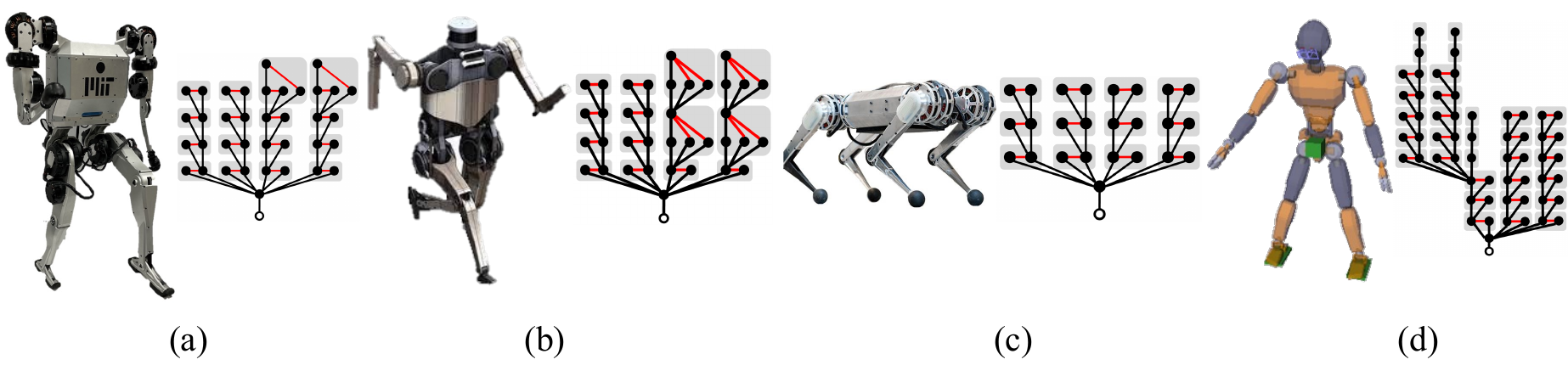}
    \caption{The proposed algorithms are benchmarked on a variety of robots, including (a) the MIT Humanoid, (b) the UIUC Tello Humanoid, (c) the MIT Mini Cheetah, and (d) the JVRC-1 Humanoid. The spanning trees and the loop joints (in red) are shown for each robot, with the grey boxes used to show clustered bodies.}
    \label{fig:robot_examples}
\end{figure*}

%% file: Sections/Helpers/NumberOfClusterFigures.tex
\begin{figure*}
    \centering
    \includegraphics[width=2\columnwidth]{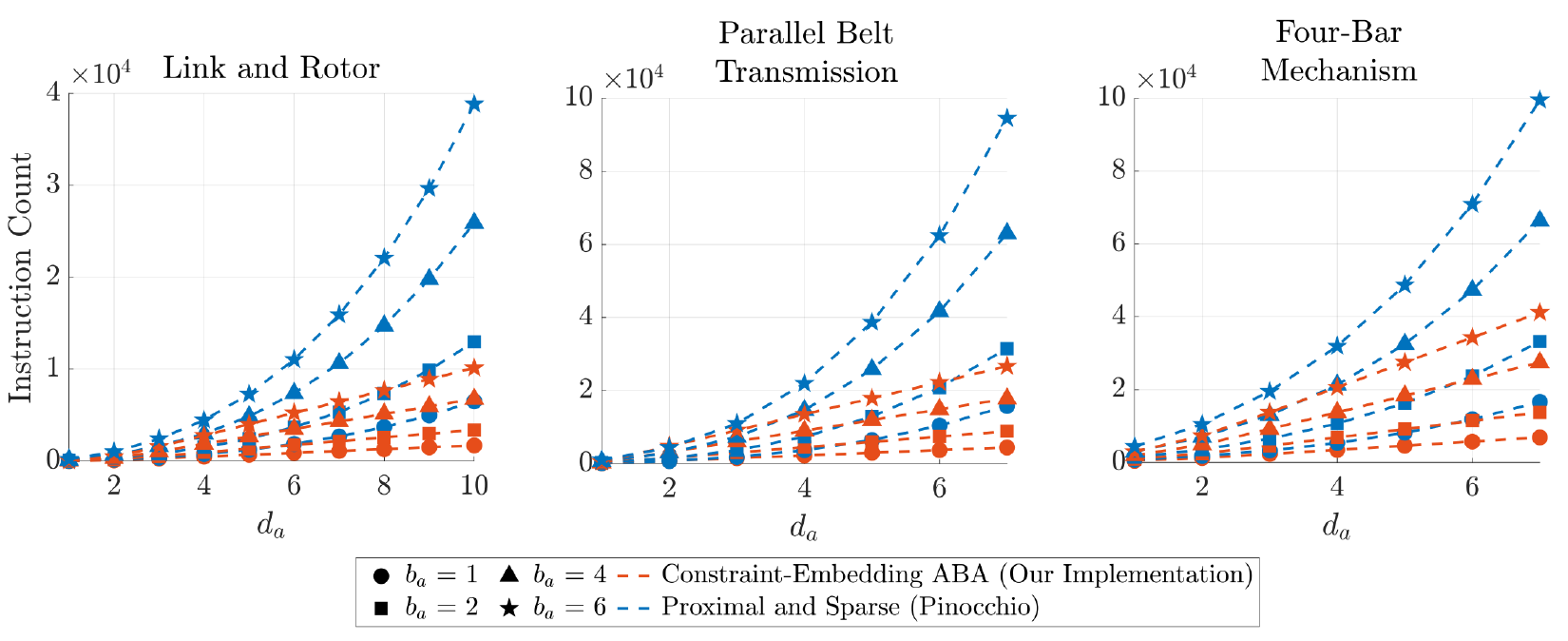}
    \caption{Effect of \aggregate~tree depth on the instruction counts of various constrained forward dynamics algorithms for the systems formed according to Fig.~\ref{fig:number_of_cluster_setup}.}
    \label{fig:number_of_cluster_instr}
\end{figure*}

\begin{figure}
    \centering
    \includegraphics[width=\columnwidth]{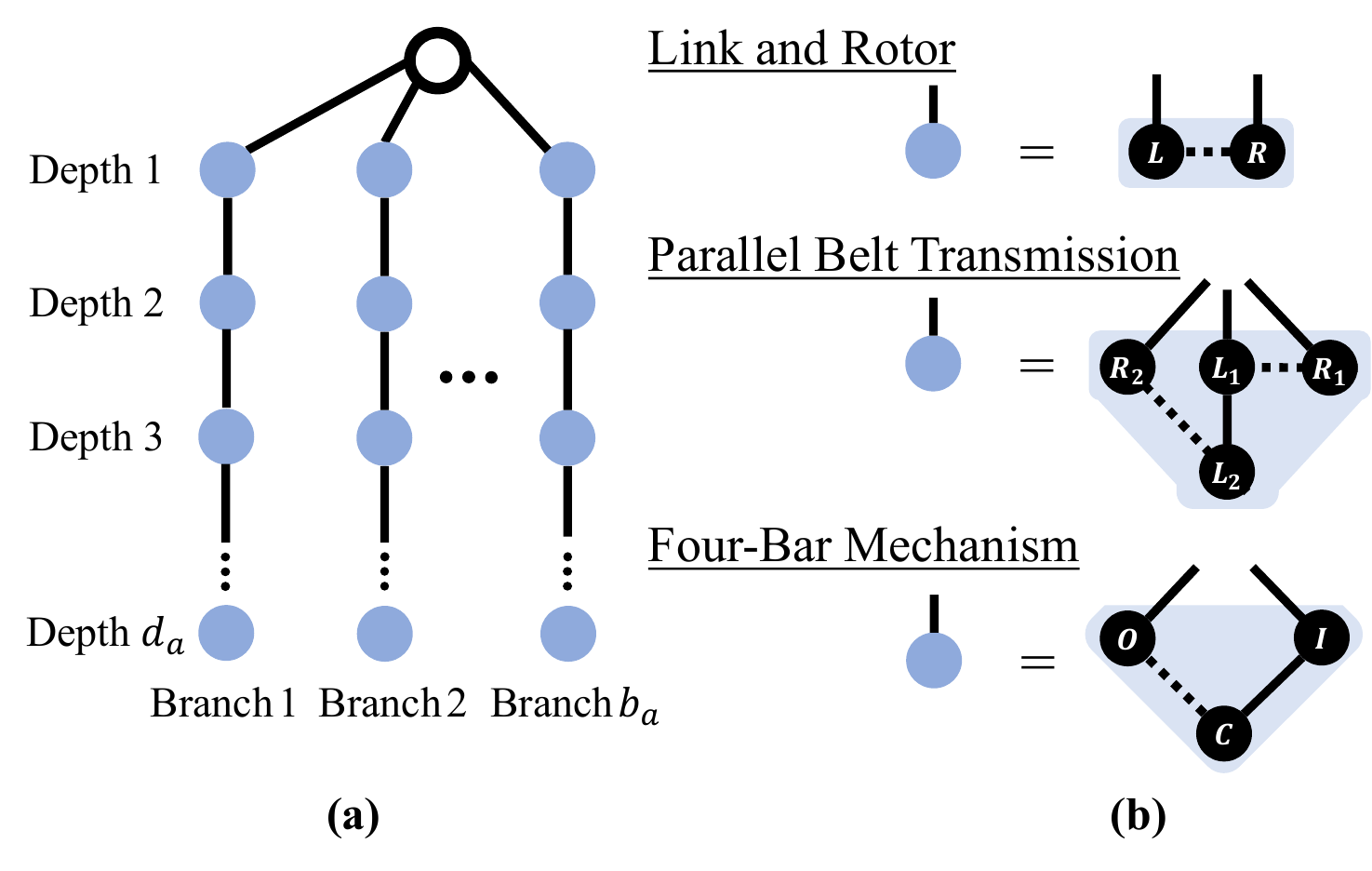}
    \caption{\textbf{(a)} \Aggregate~tree with $b_a$ branches, each having depth $d_a$. \textbf{(b)} Types of actuation sub-mechanisms and the \aggregateLink s that they create.}
    \label{fig:number_of_cluster_setup}
\end{figure}

%% file: Sections/Helpers/SizeOfClusterFigures.tex
\begin{figure*}
    \centering
    \includegraphics[width=2\columnwidth]{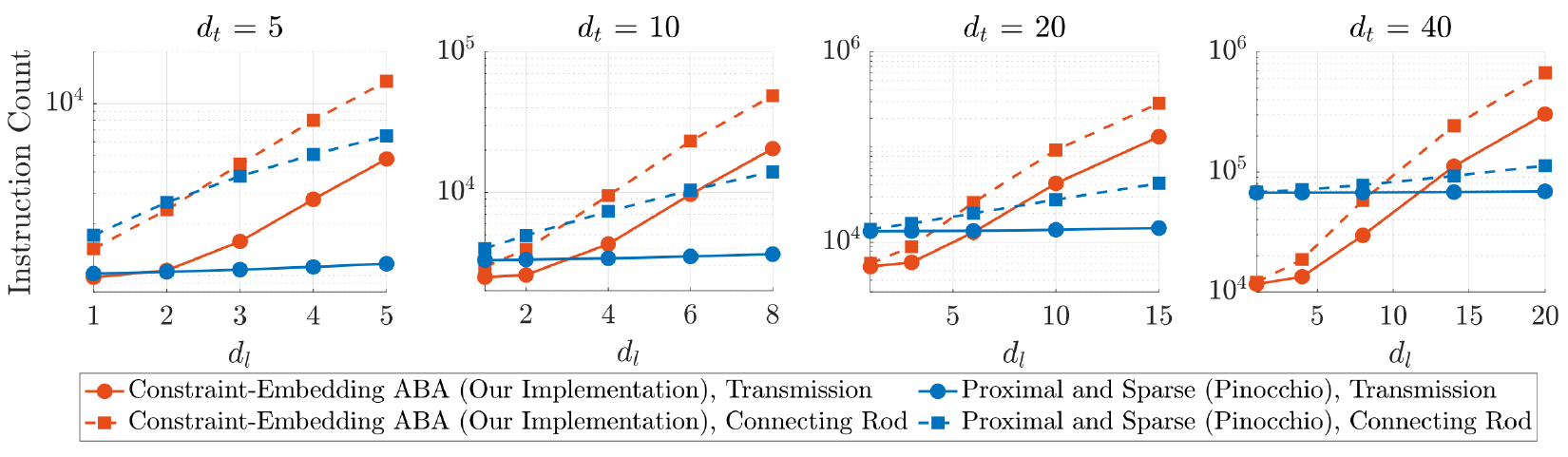}
    \caption{Effect of constraint locality on the instruction counts of various constrained forward dynamics algorithms for the systems formed according to Fig.~\ref{fig:size_of_cluster_setup}.}
    \label{fig:size_of_cluster_instr}
\end{figure*}

\begin{figure}
    \centering
    \includegraphics[width=0.8\columnwidth]{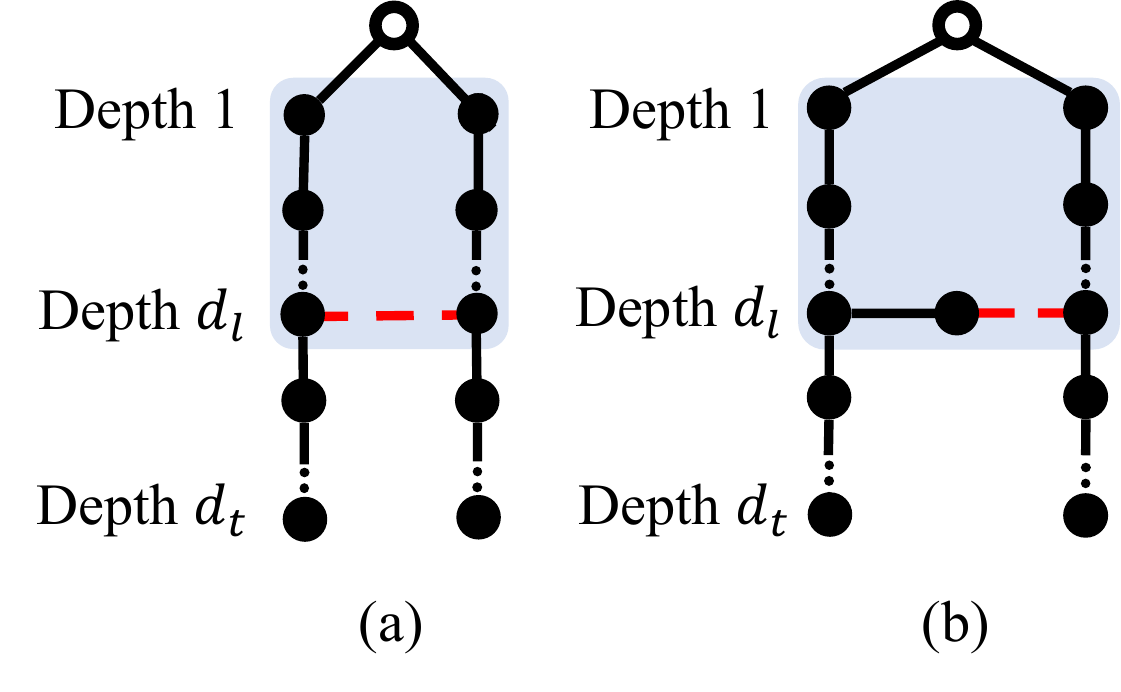}
    \caption{Spanning trees of \textbf{(a)} a transmission-constrained system and   \textbf{(b)} a connecting-rod-constrained system. Each branch has depth $d_t$, and the loop constraint is placed between the $d_l$-th body on each branch. For a given benchmark, $d_t$ is fixed and $d_l$ is increased. Dotted lines indicate continuations of bodies. Red dashed lines indicate loop constraints.}
    \label{fig:size_of_cluster_setup}
\end{figure}

%% file: Sections/Conclusion.tex
\section{Conclusion} \label{sec:conclusion}

In this paper, we have introduced a novel perspective on constraint embedding.
The perspective leverages physical analysis to generalize the concept of joint models and leads to new derivations of constraint-embedding-based recursive algorithms.
While the resulting algorithms are essentially identical to those derived with \gls{soa} techniques, the approach presented in this work is complementary to the original \gls{soa} presentation.
We hope this complementary perspective can help proliferate the use of constraint embedding in the robotics community as well as inspire new physical insights, such as the novel articulated-body assembly process presented in this work.
To that end, we also presented our open-source library of algorithms and benchmarked its contents against a state-of-the-art alternative algorithm for constrained forward dynamics.
Our benchmarking validated that constraint embedding is the most efficient approach for robots with numerous local kinematic loops.

For clarity of presentation, restrictions were placed on the number of output bodies \aggregateArticle~\aggregateLink~can have.
However, the physical analysis of constraint embedding can extend to cases of multiple output bodies and connected sub-spanning trees.
Demonstrating these extensions and considering which subclasses of them guarantee well-posedness of the multi-handle articulated inertias is left to future work.
Furthermore, in this work, we have analyzed the situations where constraint embedding or proximal and sparse approaches to constrained forward dynamics are more efficient.
However, no unified method for simultaneously applying constraint embedding to local loop constraints while leveraging proximal and sparse techniques for non-local loop constraints has been proposed to date.
Future work will also focus on this challenge, as its solution would be of widespread value to the robotics community.

%% file: Appendices/ValidatingSpanningTreeIdentities.tex
\section{Validating Spanning Tree Identities} \label{app:validate}

In Sec.~\ref{ssec:agg_force_ss}, we asserted that the \aggregate~motion and force subspace matrices must satisfy~\eqref{eqn:aggregated_dual_basis}, repeated here for convenience:
\[
    \gS_k^\top \gforceSSM^a_k = \mathbf{1}, \quad \gS_k^\top \gforceSSM^c_k = \mathbf{0}.
\]
We first demonstrate that the constructions~\eqref{eqn:aggregate_motionSSM_construction} and~\eqref{eqn:aggregate_forceSSM_construction} will always satisfy~\eqref{eqn:aggregated_dual_basis}.
First, we expand $\gS_k^\top \gforceSSM^a_k$,
\[
    \gG_k^\top\Dcat{\agglink{k}}{\S_i^\top}\gSPO{k}^\top\gSPOF{k}\Dcat{\agglink{k}}{\forceSSM^a_i}\gP_k
\]
From~\eqref{eqn:spo_sko_inv}, $\gSPO{k}^\top\gSPOF{k}=\mathbf{1}$, so we are left with
\[
\gG_k^\top\Dcat{\agglink{k}}{\S_i^\top}\Dcat{\agglink{k}}{\forceSSM^a_i}\gP_k
\]
From the rules of conventional tree joint models between individual rigid bodies~\cite{featherstone2014rigid}, $\S^\top\forceSSM^a=\mathbf{1}$, so we further reduce to $\gG_k^\top\gP_k$.
From our definition of $\gP_k$, this product must equal $\mathbf{1}$, proving that the constructions validate the first half of~\eqref{eqn:aggregated_dual_basis}.

Moving on to $\gS_k^\top \gforceSSM^c_k$, we expand it as
\[
    \gG_k^\top\Dcat{\agglink{k}}{\S_i^\top}\gSPO{k}^\top\gSPOF{k}\Dcat{\agglink{k}}{\forceSSM^c_i}\gP^c_k
\]
Repeating the $\gSPO{k}^\top\gSPOF{k}$ identity from above and again recalling the rules for individual tree joints, $\S^\top\forceSSM^c=\mathbf{0}$, leaves us with $\gG_k^\top(\mathbf{0})\gP^c_k$.
This analysis proves that the constructions validate the first half of~\eqref{eqn:aggregated_dual_basis} as well as that the choice of $\gP^c_k$ is unimportant since it is not used by any algorithms nor required to validate~\eqref{eqn:aggregated_dual_basis}.

In Chapter 3.5 of~\cite{featherstone2014rigid}, Featherstone justifies the individual tree joint relationships $\S^\top\forceSSM^a=\mathbf{1}$ and $\S^\top\forceSSM^c=\mathbf{0}$ by an analysis of the total power delivered by a tree joint.
We extend this power analysis here, showing that the total power delivered by \aggregate~joint $k$ is equal to the total power delivered by the spanning tree joints in \aggregate~joint $k$:
The power delivered by the \aggregate~joint is $<\gvJ_k,\gfJ_k>$, which from~\eqref{eqn:motion_par_rel_to_out_rel} and~\eqref{eqn:spof} can be expanded to
\[
\left(\vcat{\agglink{k}}{\v^J_i}\right)^\top\gSPO{k}^\top\gSPOF{k}\vcat{\agglink{k}}{\fj_i}. 
\]
Again using the crucial identity $\gSPO{k}^\top\gSPOF{k}=\mathbf{1}$, we are left with
\[
\left(\vcat{\agglink{k}}{\v^J_i}\right)^\top\vcat{\agglink{k}}{\fj_i},
\]
This expression can be written as the sum $\sum_{i\in\agglink{k}}\hspace{-4pt}<\fj_i,\v^J_i>$,
yielding the total power delivered by the spanning tree joints in \aggregate~joint $k$.

%% file: root.bbl
\begin{thebibliography}{10}

\bibitem{hutter2016anymal}
M.~Hutter, C.~Gehring, D.~Jud, A.~Lauber, C.~D. Bellicoso, V.~Tsounis, J.~Hwangbo, K.~Bodie, P.~Fankhauser, M.~Bloesch, {\em et~al.}, ``Anymal-a highly mobile and dynamic quadrupedal robot,'' in {\em 2016 IEEE/RSJ international conference on intelligent robots and systems (IROS)}, pp.~38--44, IEEE, 2016.

\bibitem{katz2019mini}
B.~Katz, J.~Di~Carlo, and S.~Kim, ``Mini cheetah: A platform for pushing the limits of dynamic quadruped control,'' in {\em 2019 international conference on robotics and automation (ICRA)}, pp.~6295--6301, IEEE, 2019.

\bibitem{a12023}
``A1.'' \url{https://www.unitree.com/a1/}, 2024.

\bibitem{chignoli2021humanoid}
M.~Chignoli, D.~Kim, E.~Stanger-Jones, and S.~Kim, ``The mit humanoid robot: Design, motion planning, and control for acrobatic behaviors,'' in {\em 2020 IEEE-RAS 20th International Conference on Humanoid Robots (Humanoids)}, pp.~1--8, IEEE, 2021.

\bibitem{liu2022design}
Y.~Liu, J.~Shen, J.~Zhang, X.~Zhang, T.~Zhu, and D.~Hong, ``Design and control of a miniature bipedal robot with proprioceptive actuation for dynamic behaviors,'' in {\em 2022 International Conference on Robotics and Automation (ICRA)}, pp.~8547--8553, IEEE, 2022.

\bibitem{sim2022tello}
Y.~Sim and J.~Ramos, ``Tello leg: The study of design principles and metrics for dynamic humanoid robots,'' {\em IEEE Robotics and Automation Letters}, vol.~7, no.~4, pp.~9318--9325, 2022.

\bibitem{digit2024}
``Digit.'' \url{https://robotsguide.com/robots/cassie}, 2024.

\bibitem{kangaroo2024}
``Kangaroo.'' \url{https://pal-robotics.com/robots/kangaroo/}, 2024.

\bibitem{featherstone1983calculation}
R.~Featherstone, ``The calculation of robot dynamics using articulated-body inertias,'' {\em The international journal of robotics research}, vol.~2, no.~1, pp.~13--30, 1983.

\bibitem{jain2009recursive}
A.~Jain, ``Recursive algorithms using local constraint embedding for multibody system dynamics,'' in {\em International Design Engineering Technical Conferences and Computers and Information in Engineering Conference}, vol.~49019, pp.~139--147, 2009.

\bibitem{kumar2022modular}
R.~Kumar, S.~Kumar, A.~M{\"u}ller, and F.~Kirchner, ``Modular and hybrid numerical-analytical approach-a case study on improving computational efficiency for series-parallel hybrid robots,'' in {\em 2022 IEEE/RSJ International Conference on Intelligent Robots and Systems (IROS)}, pp.~3476--3483, IEEE, 2022.

\bibitem{Luh1980}
J.~Y.~S. Luh, M.~W. Walker, and R.~P.~C. Paul, ``{On-Line Computational Scheme for Mechanical Manipulators},'' {\em Journal of Dynamic Systems, Measurement, and Control}, vol.~102, pp.~69--76, 06 1980.

\bibitem{stepanenko1976dynamics}
Y.~Stepanenko and M.~Vukobratovi{\'c}, ``Dynamics of articulated open-chain active mechanisms,'' {\em Mathematical Biosciences}, vol.~28, no.~1-2, pp.~137--170, 1976.

\bibitem{orin1979kinematic}
D.~E. Orin, R.~McGhee, M.~Vukobratovi{\'c}, and G.~Hartoch, ``Kinematic and kinetic analysis of open-chain linkages utilizing newton-euler methods,'' {\em Mathematical Biosciences}, vol.~43, no.~1-2, pp.~107--130, 1979.

\bibitem{vereshchagin1974computer}
A.~Vereshchagin, ``Computer simulation of the dynamics of complicated mechanisms of robot-manipulators,'' {\em Eng. Cybernet.}, vol.~12, pp.~65--70, 1974.

\bibitem{rodriguez1987kalman}
G.~Rodriguez, ``Kalman filtering, smoothing, and recursive robot arm forward and inverse dynamics,'' {\em IEEE Journal on Robotics and Automation}, vol.~3, no.~6, pp.~624--639, 1987.

\bibitem{rodriguez1992spatial}
G.~Rodriguez, A.~Jain, and K.~Kreutz-Delgado, ``Spatial operator algebra for multibody system dynamics,'' {\em Journal of the Astronautical Sciences}, vol.~40, no.~1, pp.~27--50, 1992.

\bibitem{featherstone2014rigid}
R.~Featherstone, {\em Rigid body dynamics algorithms}.
\newblock Springer, 2014.

\bibitem{sciavicco1995lagrange}
L.~Sciavicco, B.~Siciliano, and L.~Villani, ``Lagrange and newton-euler dynamic modeling of a gear-driven robot manipulator with inclusion of motor inertia effects,'' {\em Advanced robotics}, vol.~10, no.~3, pp.~317--334, 1995.

\bibitem{lynch2017modern}
K.~M. Lynch and F.~C. Park, {\em Modern robotics}.
\newblock Cambridge University Press, 2017.

\bibitem{becke2012extended}
M.~Becke and T.~Schlegl, ``Extended newton-euler based centrifugal/coriolis matrix factorization for geared serial robot manipulators with ideal joints,'' in {\em Proceedings of 15th International Conference MECHATRONIKA}, pp.~1--7, IEEE, 2012.

\bibitem{murphy1990recursive}
S.~H. Murphy, J.~T. Wen, and G.~N. Saridis, ``Recursive calculation of geared robot manipulator dynamics,'' in {\em Proceedings., IEEE International Conference on Robotics and Automation}, pp.~839--844, IEEE, 1990.

\bibitem{jain1990recursive}
A.~Jain and G.~Rodriguez, ``Recursive dynamics for geared robot manipulators,'' in {\em 29th IEEE Conference on Decision and Control}, pp.~1983--1988, IEEE, 1990.

\bibitem{featherstone1999divide1}
R.~Featherstone, ``A divide-and-conquer articulated-body algorithm for parallel o (log (n)) calculation of rigid-body dynamics. part 1: Basic algorithm,'' {\em The International Journal of Robotics Research}, vol.~18, no.~9, pp.~867--875, 1999.

\bibitem{featherstone1999divide2}
R.~Featherstone, ``A divide-and-conquer articulated-body algorithm for parallel o (log (n)) calculation of rigid-body dynamics. part 2: Trees, loops, and accuracy,'' {\em The International Journal of Robotics Research}, vol.~18, no.~9, pp.~876--892, 1999.

\bibitem{yamane2009comparative}
K.~Yamane and Y.~Nakamura, ``Comparative study on serial and parallel forward dynamics algorithms for kinematic chains,'' {\em The International Journal of Robotics Research}, vol.~28, no.~5, pp.~622--629, 2009.

\bibitem{sathya2023efficient}
A.~S. Sathya, H.~Bruyninckx, W.~Decr{\'e}, and G.~Pipeleers, ``Efficient constrained dynamics algorithms based on an equivalent lqr formulation using gauss' principle of least constraint,'' {\em IEEE Transactions on Robotics}, 2023.

\bibitem{popov1978manipuljacionnyje}
J.~P. Popov, A.~F. Vereshchagin, and S.~L. Zenkevich, {\em Manipuljacionnyje roboty: Dinamika i algoritmy}.
\newblock Nauka, 1978.

\bibitem{featherstone2005efficient}
R.~Featherstone, ``Efficient factorization of the joint-space inertia matrix for branched kinematic trees,'' {\em The International Journal of Robotics Research}, vol.~24, no.~6, pp.~487--500, 2005.

\bibitem{carpentier2021proximal}
J.~Carpentier, R.~Budhiraja, and N.~Mansard, ``Proximal and sparse resolution of constrained dynamic equations,'' in {\em Robotics: Science and Systems 2021}, 2021.

\bibitem{brandl1987algorithm}
H.~Brandl, ``An algorithm for the simulation of multibody systems with kinematic loops,'' in {\em Proc. the IFToMM Seventh World Congress on the Theory of Machines and Mechanisms}, Sevilla, 1987.

\bibitem{critchley2003generalized}
J.~Critchley and K.~S. Anderson, ``A generalized recursive coordinate reduction method for multibody system dynamics,'' {\em International Journal for Multiscale Computational Engineering}, vol.~1, no.~2\&3, 2003.

\bibitem{muller2022constraint}
A.~M{\"u}ller, ``A constraint embedding approach for dynamics modeling of parallel kinematic manipulators with hybrid limbs,'' {\em Robotics and Autonomous Systems}, p.~104187, 2022.

\bibitem{jain2012multibody_part1}
A.~Jain, ``Multibody graph transformations and analysis, part i: Tree topology systems,'' {\em Nonlinear dynamics}, vol.~67, no.~4, pp.~2779--2797, 2012.

\bibitem{jain2012multibody_part2}
A.~Jain, ``Multibody graph transformations and analysis, part ii: Closed-chain constraint embedding,'' {\em Nonlinear dynamics}, vol.~67, no.~4, p.~2153–2170, 2012.

\bibitem{carpentier2019pinocchio}
J.~Carpentier, G.~Saurel, G.~Buondonno, J.~Mirabel, F.~Lamiraux, O.~Stasse, and N.~Mansard, ``The pinocchio c++ library: A fast and flexible implementation of rigid body dynamics algorithms and their analytical derivatives,'' in {\em 2019 IEEE/SICE International Symposium on System Integration (SII)}, pp.~614--619, IEEE, 2019.

\bibitem{jain2011graph1}
A.~Jain, ``Graph theoretic foundations of multibody dynamics: Part i: structural properties,'' {\em Multibody system dynamics}, vol.~26, pp.~307--333, 2011.

\bibitem{featherstone2010beginner}
R.~Featherstone, ``A beginner's guide to 6-d vectors (part 1),'' {\em IEEE robotics \& automation magazine}, vol.~17, no.~3, pp.~83--94, 2010.

\bibitem{featherstone2010beginner2}
R.~Featherstone, ``A beginner's guide to 6-d vectors (part 2)[tutorial],'' {\em IEEE robotics \& automation magazine}, vol.~17, no.~4, pp.~88--99, 2010.

\bibitem{jain2012efficient}
A.~Jain, C.~Crean, C.~Kuo, and M.~B. Quadrelli, ``Efficient constraint modeling for closed-chain dynamics,'' in {\em The 2nd Joint International Conference on Multibody System Dynamics, Stuttgart, Germany}, Citeseer, 2012.

\bibitem{kumar2020analytical}
S.~Kumar, K.~A.~v. Szadkowski, A.~Mueller, and F.~Kirchner, ``An analytical and modular software workbench for solving kinematics and dynamics of series-parallel hybrid robots,'' {\em Journal of Mechanisms and Robotics}, vol.~12, no.~2, 2020.

\bibitem{gauss1829neues}
C.~F. Gau{\ss}, ``{\"U}ber ein neues allgemeines grundgesetz der mechanik.,'' 1829.

\bibitem{grbda2023}
{ROAM Lab}, ``Generalized rbda.'' \url{https://github.com/ROAM-Lab-ND/generalized_rbda}, 2023.
\newblock Branch: v2.1.0.

\bibitem{andersson2019casadi}
J.~A. Andersson, J.~Gillis, G.~Horn, J.~B. Rawlings, and M.~Diehl, ``Casadi: a software framework for nonlinear optimization and optimal control,'' {\em Mathematical Programming Computation}, vol.~11, pp.~1--36, 2019.

\bibitem{wensing2012reduced}
P.~Wensing, R.~Featherstone, and D.~E. Orin, ``A reduced-order recursive algorithm for the computation of the operational-space inertia matrix,'' in {\em 2012 IEEE International Conference on Robotics and Automation}, pp.~4911--4917, IEEE, 2012.

\bibitem{todorov2012mujoco}
E.~Todorov, T.~Erez, and Y.~Tassa, ``Mujoco: A physics engine for model-based control,'' in {\em 2012 IEEE/RSJ International Conference on Intelligent Robots and Systems}, pp.~5026--5033, IEEE, 2012.

\bibitem{peng2018sim}
X.~B. Peng, M.~Andrychowicz, W.~Zaremba, and P.~Abbeel, ``Sim-to-real transfer of robotic control with dynamics randomization,'' in {\em 2018 IEEE international conference on robotics and automation (ICRA)}, pp.~3803--3810, IEEE, 2018.

\bibitem{okugawa2015proposal}
M.~Okugawa, K.~Oogane, M.~Shimizu, Y.~Ohtsubo, T.~Kimura, T.~Takahashi, and S.~Tadokoro, ``Proposal of inspection and rescue tasks for tunnel disasters—task development of japan virtual robotics challenge,'' in {\em 2015 IEEE international symposium on safety, security, and rescue robotics (SSRR)}, pp.~1--2, IEEE, 2015.

\end{thebibliography}
